\documentclass{ecai}
\usepackage[ruled]{algorithm2e} 
\usepackage{booktabs} 
\usepackage{graphicx}
\usepackage{latexsym}
\usepackage{amsmath, amssymb}
\usepackage{bm}
\usepackage[table]{xcolor}
\definecolor{lightgray}{RGB}{211 ,211 ,211}
\definecolor{lightred}{RGB}{235 ,113 ,106}
\definecolor{lightblue}{RGB}{120 ,167 ,213}
\definecolor{lightyellow}{RGB}{247 ,218 ,119}
\usepackage{graphicx}
\usepackage{subcaption} 
\usepackage{tikz} 
\usepackage{natbib} 
\usepackage{amssymb}
\usepackage{pifont}
\usepackage{natbib}
\setcitestyle{numbers,square}
\usepackage[colorlinks=true, linkcolor=blue, citecolor=blue, urlcolor=cyan]{hyperref}

\paperid{1762}        

\begin{document}

\begin{frontmatter}

\title{SETTP: \underline{S}tyle \underline{E}xtraction and \underline{T}unable Inference via Dual-level \underline{T}ransferable \underline{P}rompt Learning}

\author[A;B;**]{\fnms{Chunzhen}~\snm{Jin}\footnote{Equal contribution.}}
\author[C;**]{\fnms{Yongfeng}~\snm{Huang}\footnotemark}
\author[A;B]{\fnms{Yaqi}~\snm{Wang}}
\author[A;B;D]{\fnms{Peng}~\snm{Cao}\thanks{Corresponding Author. Email: caopeng@mail.neu.edu.cn.}}
\author[E]{\fnms{Osmar}~\snm{Zaïane}}

\address[A]{Computer Science and Engineering, Northeastern University, Shenyang, China}
\address[B]{Key Laboratory of Intelligent Computing in Medical Image of Ministry of Education, Northeastern University, Shenyang, China}
\address[C]{Department of Computer Science and Engineering, The Chinese University of Hong Kong}
\address[D]{National Frontiers Science Center for Industrial Intelligence and Systems Optimization, Shenyang, China}
\address[E]{Amii, University of Alberta, Edmonton, Alberta, Canada}

\begin{abstract}
Text style transfer, an important research direction in natural language processing, aims to adapt the text to various preferences but often faces challenges with limited resources. 
In this work, we introduce a novel method termed \underline{S}tyle \underline{E}xtraction and \underline{T}unable Inference via Dual-level \underline{T}ransferable \underline{P}rompt Learning (SETTP) for effective style transfer in low-resource scenarios.
First, SETTP learns source style-level prompts containing fundamental style characteristics from high-resource style transfer. 
During training, the source style-level prompts are transferred through an attention module to derive a target style-level prompt for beneficial knowledge provision in low-resource style transfer.  
Additionally, we propose instance-level prompts obtained by clustering the target resources based on the semantic content to reduce semantic bias.
We also propose an automated evaluation approach of style similarity based on alignment with human evaluations using ChatGPT-4.
Our experiments across three resourceful styles show that SETTP requires only 1/20th of the data volume to achieve performance comparable to state-of-the-art (SOTA) methods. 
In tasks involving scarce data like writing style and role style, SETTP outperforms previous methods by 16.24\%.

\end{abstract}

\end{frontmatter}

\section{Introduction}
Text style transfer (TST) is a critical and challenging task in natural language generation. It aims to endow the text with a new style without altering its fundamental meaning \cite{kn:gao22}. Recently, pre-trained language models (PLMs) have demonstrated significant potential in TST  \cite{openai2023gpt4} \cite{touvron2023llama}. 
Numerous studies concentrated on TST with abundant resources, such as sentiment transfer \cite{santos2018fighting} or format transfer \cite{li2018delete}. They have achieved notable results through fine-tuning with datasets reaching hundreds of thousands of entries.
In real-world scenarios, it is inevitable to confront styles with only limited labeled data, particularly in the context of writing style transfer \cite{zhu2023storytrans} and specific role style transfer \cite{xu2023specializing}. Conventional methods via fine-tuning with scarce data frequently failed to yield satisfactory results. A common approach involves leveraging large-scale datasets from different domains, such as the transferable prompts method \cite{li2022learning}. This technique utilizes the source prompts generated from the source tasks to transfer knowledge to the target task, successfully alleviating data scarcity. 
\begin{figure}[!htb]

  \begin{subfigure}{.23\textwidth}
    \centering
    \includegraphics[width=\linewidth]{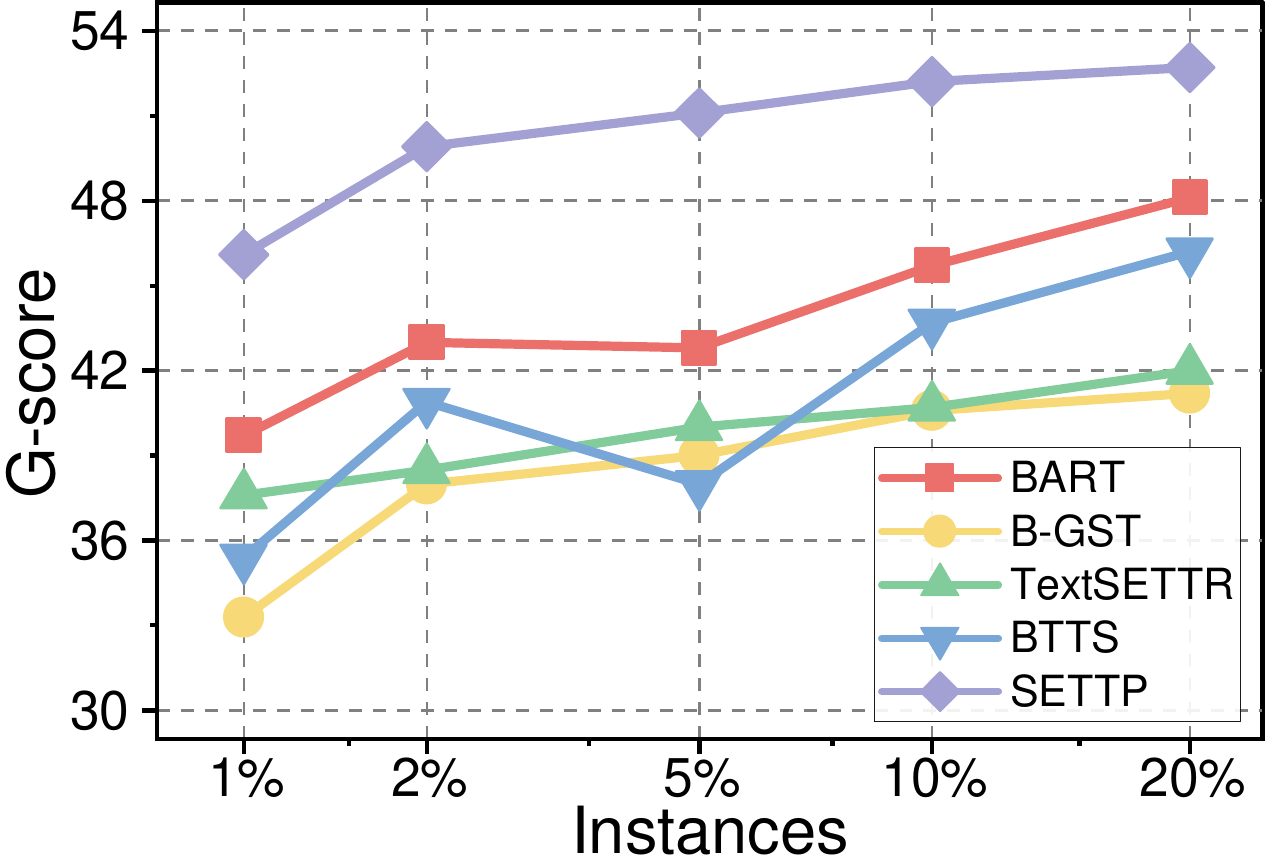}
\caption{Genshin Dataset}
\label{fig:instance1}
  \end{subfigure}%
  \hfill
  \begin{subfigure}{.23\textwidth}
    \centering
    \includegraphics[width=\linewidth]{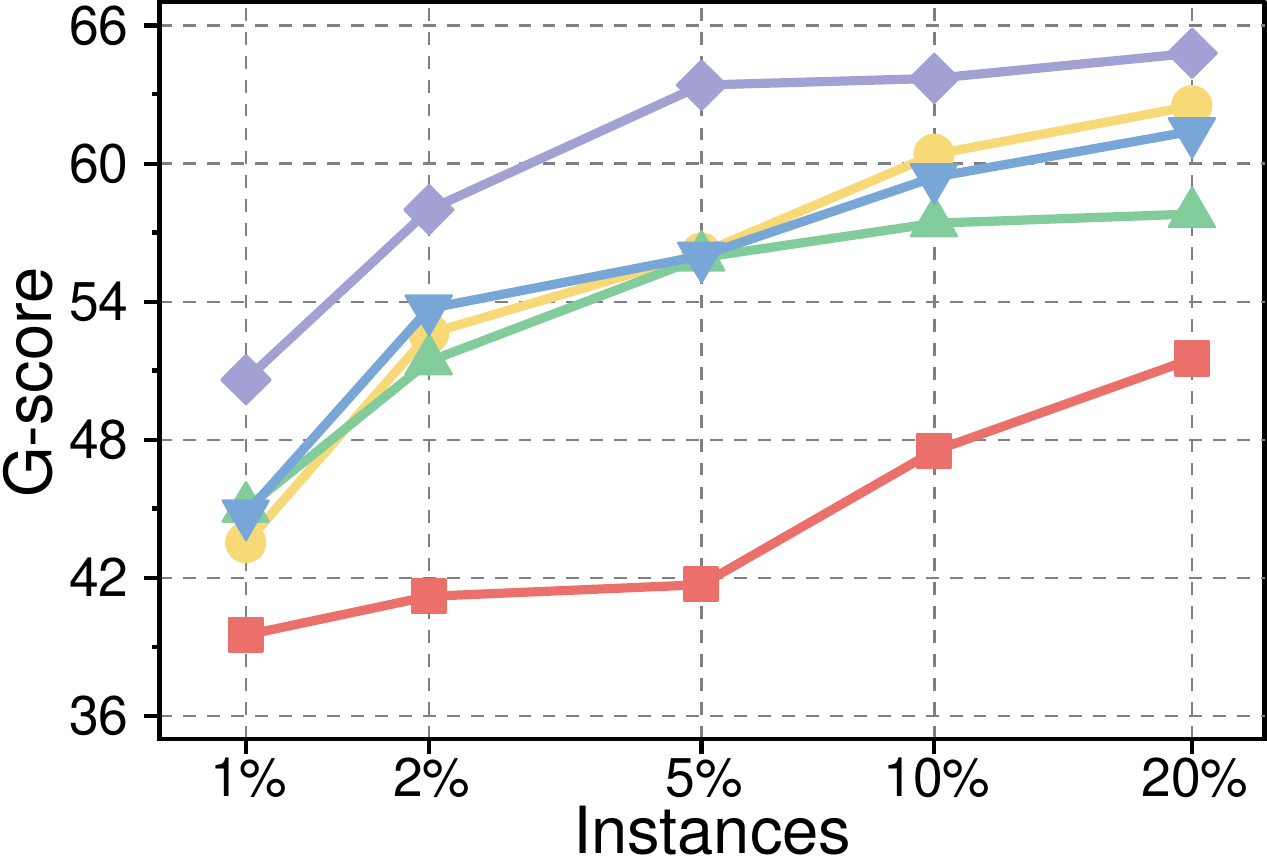}
    \caption{GYAFC E\&M Dataset}
    \label{fig:instance2}
  \end{subfigure}%

\caption{Comparison under various shots of instances.}\label{few-fig}
\end{figure}

However, applying transferable prompts method to the TST field presents two main challenges. \textbf{Challenge 1}: The prompts from the source TST tasks are highly specialized, making it challenging to transfer the knowledge to new TST tasks effectively.
\textbf{Challenge 2}: For a single TST task, even a well-learned prompt may not fit all data in a large population, leading to semantic bias \cite{kim2023size}. Devising an adaptive prompt transfer strategy to mitigate this bias is non-trivial.

This work introduces a novel method named \textbf{\textit{Style Extraction and Tunable Inference via Dual-level Transferable Prompt Learning} (SETTP)} to overcome the aforementioned challenges. We first learn some transferable soft prompts in high-resource style transfer tasks called  \textbf{Source Style-level Prompts}. These prompts encapsulate general representations beneficial to low-resource styles and are stored in the style-level Prompt Pool for subsequent reuse. Then, valuable parts of the source style-level prompts are integrated to derive a \textbf{Target Style-level Prompt} through the Adaptive Attentional Retrieval (AAR) technique, addressing Challenge 1. AAR learns a set of Key vectors for style-level prompts and uses the average embedding of the target style as the Query vector to derive target style-level prompts (Value), which are used to transfer knowledge. Additionally, considering the transferable comprehensive representation might not be fully adaptable to all data in a low-resource style,  we design Prompt-tunable Inference and \textbf{Instance-level Prompts}, addressing Challenge 2. In advance, we cluster the target low-resource data according to content, aiming to handle semantically complex scenarios and train a prompt for each cluster. These prompts are called instance-level prompts and contain specific style representation modes based on semantics. The most content-relevant prompts will be embedded during inference to enhance semantic alignment.

Another \textbf{Challenge} is Inappropriate Evaluation Metrics. Recent studies \cite{chen2023exploring} \cite{ostheimer2023standardization} highlight that the correlation between automated metrics and human evaluations is notably weak. In response, we propose an evaluation method that utilizes \textsc{\textbf{ChatGPT-4}} \cite{openai2023gpt4}, which more closely aligns with human judgments concerning style similarity.



SETTP offers three \textbf{Key Advantages} over previous TST methods: First, Figure~\ref{few-fig} demonstrates SETTP has strong few-shot learning capabilities and outperforms the strong baseline by 16.24\% in extreme low-resource scenarios. Second, the prompt pool is a modular component that allows for the easy addition, reuse, and removal of soft prompts, sharing knowledge across different tasks. Third, whereas previous methods designed additional experience-dependent regularization constraints to reduce semantic bias, SETTP is more flexible and adjusts stylistic expressions based on Prompt-tunable Inference.


Our contributions are as follows: 
(1) We propose a TST method based on transferable prompts to address the challenges of low-resource style transfer. 
(2) Our method achieve the best average performance in thirteen TST tasks compared with ten strong baselines.
(3) We propose an automatic evaluation method based on ChatGPT-4 and demonstrate its high consistency with human evaluations regarding style similarity.Our code is available~\footnote{https://github.com/Kimo1116/tst}.

\section{Preliminary}
\label{Knowledge}
\subsection{Formulation of Text Style Transfer}


Natural language generation (NLG) is designed to model the conditional probability $\Pr(y|x)$, where $x = \{\mathbf{w}_1, ..., \mathbf{w}_{t_x}\}$ represents the input text in the source style, and $y = \{\mathbf{z}_1, ..., \mathbf{z}_{t_y}\}$ denotes the output text in the target style. Formally, $x$ and $y$ are sequences of tokens from a vocabulary $\mathcal{V}$, while $t_x$ and $t_y$ are the number of tokens. 

Text Style Transfer (TST), as a specialized task within NLG, aims to change the style of $x$ to produce $y$, which keeps the original meaning but adopts a new style. We use prompt techniques to achieve this, adding extra stylistic details to PLMs during generation. Unlike traditional methods that attach fixed prompts to the input, we use learnable continuous (soft) prompts. For an input of $t_x$ tokens $x = \{\mathbf{w}_1, ..., \mathbf{w}_{t_x}\}$ , we embed it using a PLM into an embedding matrix $E_x \in \mathbb{R}^{t_x \times e}$, where $e$ is the dimension of the embedding space. Our style prompt $P$ is represented as a matrix $E_P \in \mathbb{R}^{m \times e}$, where $m$ is the number of prompt vectors. Then, $P$ is combined with the embedded input to create a unified matrix $[E_P; E_x] \in \mathbb{R}^{(m+t_x)\times e}$. The PLM processes this matrix as a standard sequence, aiming to maximize the likelihood of generating the stylistically transformed ground truth y, i.e., $\Pr(y|[P; x])$.

\subsection{Transfer Learning and Prompt Learning}
\textbf{Transfer learning} is a machine learning strategy where a model $f_{\theta_S}(\cdot)$ is first trained on a source task $S$ under the learnable parameter $\theta_S$. $\theta_S$ can effectively capture feature expressions beneficial to the source task. Subsequently, $f_{\theta_S}(\cdot)$ is transferred to a target task $T$. A fine-tuning process updates the model parameters to $\theta_T$ to effectively adapt to the target task with the initial parameter $\theta_S$. 

\textbf{Prompt learning} guides PLMs in exhibiting output on specific tasks by designing specific prompts. With a given model $f(\cdot)$ and corresponding parameter $\theta_{S}$, we only need to adjust a specific prompt $P_{T}$ to achieve an approximate equivalence between $f_{\theta_T}(\cdot)$ and $f_{\theta_S}(\cdot; P_T)$, thereby effectively enhancing the model's performance on the target task $T$.

\subsection{Transferable Prompt Learning for TST}
\label{ptl}
Considering a set of source style transfer tasks $S = \{S_1, ..., S_N\}$, each task $S_n = \{(x_i^n, y_i^n)\}^{k_n}_{i=1}$ comprises $k_n$ tuples of input text $x_i^n$ in the source style and its corresponding output text $y_i^n$ in the target style.
The objective for a new style transfer task $T$ is to leverage the stylistic knowledge extracted from the source tasks $S$ to enhance the model's performance in rendering the target style. In this setting, the parameters of the underlying model remain unchanged, and the style transfer tasks are facilitated by appending soft prompts to the input. Specifically, we will learn a distinct source style prompt $P_n$ for each source task $S_n$, based on a shared, unmodified model $f_{\theta_S}(\cdot; P)$, by maximizing the likelihood $\Pr({y_i^n}|[{P_n}; {x_i^n}])$. The key innovation lies in transferring $P_n$ to a new style transfer task via interpolated target prompt $P_T$, enabling the target task to be executed effectively in low-resource scenarios. Ultimately, our objective is to ensure that the initial model equipped with a knowledgeable prompt $f_{\theta_S}(\cdot; P_T)$ outperforms $f_{\theta_T}(\cdot)$ after optimizing $\theta_S$ to $\theta_T$ in environments characterized by limited data availability.

\begin{figure*}[!htb]
    \centering
    \includegraphics[width=0.85\textwidth]{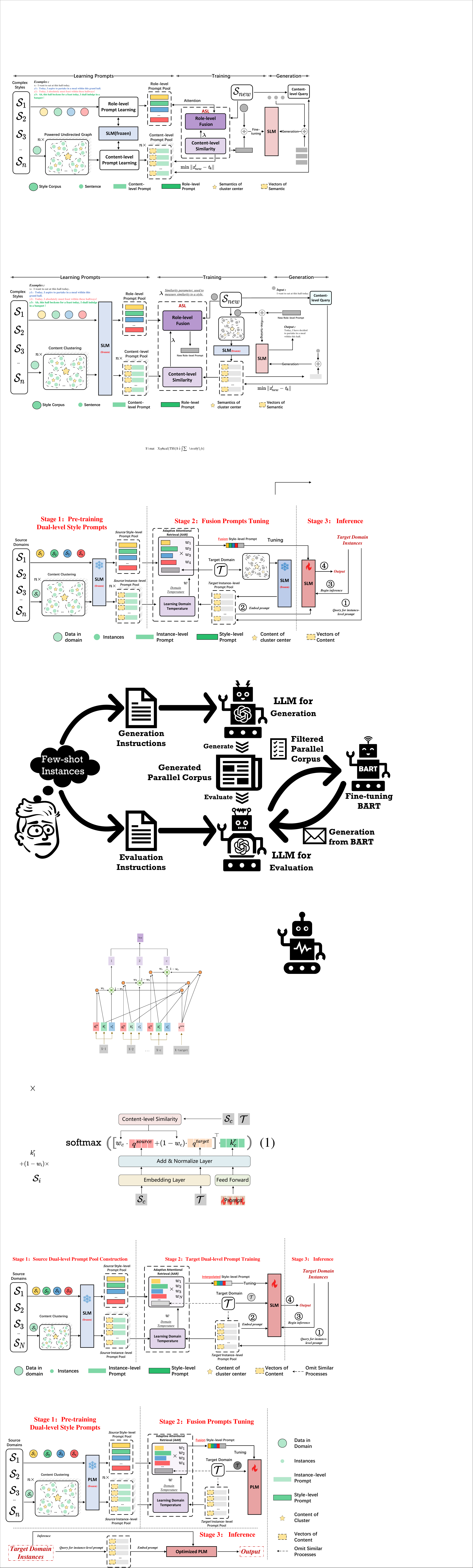}
    \caption{The overall architecture of the proposed SETTP. In \textit{stage 1}, we adopt a frozen PLM (e.g., BART) to extract both style-level and instance-level prompts from source domain $\mathcal{S}$. In \textit{stage 2}, based on the constructed prompt pools, we embed style-level prompts for the target domain $\mathcal{T}$ by fusing source style-level information via \textit{AAR}. In \textit{stage 3}, we embed the target instance-level prompt for each sentence during inference by calculating the L2 distance. \ding{172}\ding{173}\ding{174}\ding{175} represents the order of the inference process.}
\label{framwork}
\end{figure*}

\section{Methodology}
\subsection{Overview}
As shown in Figure~\ref{framwork}, the proposed method \textbf{\textit{Style Extraction and Tunable Inference via Dual-level Transferable Prompt Learning} (SETTP)} consists of three stages: 

\textit{Stage 1: Source Dual-level Prompt Pool Construction}  involves learning and storing two types of prompts through a frozen model. Style-level prompts are designed to capture the overarching style of the source domain, which encapsulates comprehensive knowledge and serves as the primary representation for prompt interpolation in the next stage. 
Instance-level prompts are obtained through content-based clustering, containing reinforced information about the coupling between content and style, thereby reducing semantic bias.

\textit{Stage 2: Target Dual-level Prompt Training} utilizes techniques such as AAR and Domain Temperature constraints to interpolate the previously isolated style-level prompt effectively. The attention mechanism selectively focuses on relevant aspects of styles, while Domain Temperature constraints regulate the influence of each domain, ensuring a balanced and effective interpolation. For the target instance-level prompt, the training process is the same as the source.

\textit{Stage 3: Adapting to Target Domain Styles with Prompt-Tunable Inference } is the phase where the frozen model parameters are tuned by leveraging the most relevant prompts for specific instances. This stage is characterized by a strategic retrieval of prompts most applicable to the instance. Drawing on the rich pool of instance-level prompts reduces semantic bias during inference. 


\subsection{Source Dual-level Prompt Pool Construction}
\label{Stage 1}
In the first stage, we obtain transferable knowledge in the source domain through prompt learning like Section~\ref{ptl}. Learned source style-level prompt $P_S^{style}$ and source instance-level prompt $P_S^{ins}$ are stored in style-level and instance-level prompt pools, respectively.

\subsubsection{Source Style-level Prompt Pre-training}
We learn Style-level Prompts by freezing the model parameters. Specifically, the style-level prompt $ P_S^{style}$ is randomly initialized as a parameter vector. Then, optimal $ P_S^{style}$ is obtained by maximizing the likelihood of the ground truth. Given a supervised dataset $(\mathcal{X}, \mathcal{Y})_n$ for the specific style transfer task $S_n$, the loss of pre-training the style-level prompt is defined as follows:
\begin{equation}
 \mathcal{L}_{\text{pre-style}}(\theta_S) = -\underset{{(x, y) \sim (\mathcal{X}, \mathcal{Y})_n}}{\mathbb{E}}  \left[\log \Pr\left(y | [P_S^{style}; x]; \theta_S \right)\right]  \
 \label{style_training}
\end{equation}
During training, the parameter $\theta_S$ of the overall model is frozen, and $P_S^{style}$ can be optimized. This process will be repeated $N$ times until all source style transfer tasks have corresponding $P_S^{style}$.

\subsubsection{Source Instance-level Prompt Pre-training}
Instance-level Prompts are designed to generate fine-grained prompts for input instances with similar content, thus necessitating data pre-processing. Through the spectral clustering algorithm, we categorize the inputs of the source corpus into several content clusters $\mathcal{C} \in \mathcal{G}$, where $\mathcal{G}$ is a weighted undirected graph. Specifically, each input is treated as a node $p$, and the weight between node $p_i$ and $p_j$ is $\mathbb{W}_{i,j} = 1/(1 + \|p_i - p_j\|)$. To ensure each cluster has sufficient nodes, we employ the min-max cut strategy to segment $\mathcal{G}$, resulting in $\mathcal{C} = \{\mathcal{C}_1, \mathcal{C}_2, \ldots, \mathcal{C}_L\}$, where $L$ is the number of clusters.

Subsequently, we train a specific instance-level prompt for each $C_k$. 
The training process of the specific style transfer task $S_n$ is similar to Equations \ref{style_training}. Formally, the loss is defined as follows:
\begin{equation}
 \mathcal{L}_{\text{pre-ins}}(\theta_S) =-\sum\limits_{k=1}^{L}
{\mathbb{E}}_{{(x, y) \sim (\mathcal{X}, \mathcal{Y})_n^{k}}}  \left[\log \Pr\left(y | [P_k^{ins}; x]; \theta_S \right)\right]  \
\end{equation}
where $\theta_S$ is frozen and shared among all tasks. After training,  we obtain prompts set $ P_n^{ins} = \{P_1^{ins}, P_2^{ins},  \ldots, P_L^{ins}\} $, where each prompt aligns with different semantics.


\subsubsection{Structure and Retrieval of Prompt Pool}
Given the source style-level prompts $P_S^{style}$ and a set of source instance-level prompts $P_S^{ins}$ for each domain, we construct two Prompt Pool $\mathbf{P}^{style}$ and $\mathbf{P}^{ins}$ to store these prompts for facilitating the reuse and retrieval respectively. From a technical implementation perspective, the prompt pools are two types of memory networks: \textit{Key-Value} memory network \cite{miller2016keyvalue} and \textit{End-to-End} \cite{weston2015memory} memory network. Both can learn how to retrieve relevant information from a vast amount of stored data. The storage structures are as follows:
\begin{eqnarray}
&\mathbf{P}^{style} = \{ P^{style}_n;\bm{k}^n\}_{n=1}^N  \\
&\mathbf{P}^{ins} = \{ (P_l^{ins};\bm{t}^l)_{l=1}^{L_n}\}_{n=1}^N
\end{eqnarray}
where $\bm{k}$ is a learnable key vector of the style-level prompt and $\bm{t} \in \mathbb{R}^{e}$ is the embedded content vector of the clustering center.

The retrieval methods are different: In $\mathbf{P}^{style}$, the stored $P_S^{style}$ can serve as a basis for representation, which can be derived through key-value queries in the next stage; In $\mathbf{P}^{ins}$, we perform $L_2$ distance searches between target content vector $\bm{\tilde{t}}$ and stored content vector $\bm{t}$. Searches require additional computing resources and pre-clustering can reduce the amount of computing.


\subsection{Target Dual-level Prompt Training}
In the target TST task, we first employ similar methods described in \textit{Stage 1} to generate \textit{\textbf{Preliminary}} Target Style-level Prompt and Target Instance-level Prompts. Subsequently, Section~\ref{Attentional} retrieves useful parts of source style-level prompts for interpolating the \textbf{\textit{Final}} Target Style-level Prompt in Section~\ref{Interpolation}.

\subsubsection{Attentional Retrieval of Source Prompt}
\label{Attentional}
SETTP introduces the computation of attention between input prompts to control the influence of source prompts on the target prompt. We name this module \textbf{\textit{Adaptive Attentional Retrieval (AAR)}}. Specifically, we assign a learnable \textit{Key} vector $\bm{k} \in \mathbb{R}^{d}$ to each style-level prompt $P^{style}$, where $d$ is the \textit{Key} embedding size. The style-level prompt vectors are the initialization of the \textit{Key}, further optimized with gradient propagation. We then average each style's content embedding to obtain \textit{Query} vectors for both source-domain and target-domain styles, denoted as $\bm{q}_S \in \mathbb{R}^{d}$ and $\bm{q}_T \in \mathbb{R}^{d}$. This conversion leverages the capabilities of BERT \cite{devlin2019bert}. For a source style $\mathcal{S}_n$, its Retrieval Score $s_n$ is calculated as follows:
\begin{equation}
s_n = \text{softmax}\left( \left[ w_n \cdot \bm{q}^n_S + (1 - w_n) \cdot \bm{q}_T \right]^{\top} \cdot \bm{k}^n_S \right)
\end{equation}
where $w_n$ represents the parameter \textit{\textbf{Domain Temperature}}, which balances the learning levels between high and low-resource domains. Acknowledging that knowledge acquired from high-resource domains is generally perceived as more convincing \cite{goswami2023switchprompt}, this parameter is designed to maintain equilibrium by addressing the potential biases low-resource settings might cause. The specific learning process is detailed in Appendix Algorithm~\ref{al}.
\label{DT}

\subsubsection{Target Prompt Interpolation}
\label{Interpolation}
After obtaining the preliminary target style-level prompt $P_T^{style}$and retrieval scores $s_n$ of all source style-level prompts $P_S^{style}$, the final target style-level prompt $\tilde{P_T^{style}}$ is computed as follows:
\begin{equation}
\tilde{P_T^{style}}=  P_T^{style} + \lambda \cdot\sum_{n=1}^{N} s_n \cdot P_S^{style} 
\end{equation}
where $N$ is the number of source domain, $\lambda$ is a hyperparameter. 

\subsubsection{Interpolated Target Prompt Tuning}
The interpolated final target style-level prompt $\tilde{P_T^{style}}$ contains the overall style characteristics of the target domain and supplementary knowledge from the source domain. Subsequently, $\tilde{P_T^{style}}$ is added to the embedding layer, enabling the model $f(\theta; \tilde{P_T^{style}})$ to acquire knowledge for more efficient training initiation. $\theta_T$ is optimized in the target domain using Maximum Likelihood Estimation as follows:
\begin{equation}
\mathcal{L}_{\text{style}}(\theta_T) = -\underset{(x, y) \sim (\mathcal{X}, \mathcal{Y})_T}{\mathbb{E}}\left[\log \Pr\left(y \mid [\tilde{P_T^{style}},x]; \theta_T \right)\right]
\end{equation}
where $\theta_T$ is updated via gradient descent, while $P_S^{style}$ and $\tilde{P_T^{style}}$ are untouched to preserve the encoded style-related knowledge.

\subsection{Adapting to Target Domain Styles with Prompt-Tunable Inference }
In preliminary experiments, we found that style-level prompts containing style-invariant representations might not fully adapt to the target domain. The model exhibits overly high semantics bias and insufficiently clear style variation. To address this issue, we introduce \textit{Prompt-Tunable Inference}.

\subsubsection{Query and Embed Instance-level Prompts  (\ding{172} \ding{173})}
For any given input instance, the first step involves transforming the content into an embedded vector $\bm{t}_i \in \mathbb{R}^{e}$, which is aligned with content vectors in the instance-level prompt pool. 

Following the encoding, the process matches the encoded query about $\bm{t}_i$ against a pre-defined set of instance-level prompts. These prompts are organized around cluster centers within an instance-level prompt pool, which essentially acts as a repository of content-related prompts designed to guide the model's response generation. The matching criterion relies on the $L_2$ distance, a measure of the Euclidean distance between vectors, to identify the matched prompt $\bm{t}_l$ that minimizes this distance to $\bm{t}_i$, indicating the highest content similarity.
Mathematically, this retrieval process is as follows:
\begin{equation}
P_T^{ins}= \mathbf{P}^{ins}\{\tilde{P_l^{ins}};(\bm{t}_l|min\|\bm{t}_i-\bm{t}_l\|)\}
\end{equation}
Where $\tilde{P_l^{ins}}$ represents the instance-level prompt corresponding to $\bm{t}_l$ that satisfies $\|\bm{t}_i-\bm{t}_l\|$. This expression formalizes the selection of the most relevant instance-level prompts $P_T^{ins}$ based on the input instance.

\subsubsection{Tunable Inference (\ding{174} \ding{175})}
Once $P_T^{ins}$ is identified, the next step involves generating a response. The generation process is grounded in the principles of language modeling, explicitly leveraging the probability distribution $\Pr(y|P_T^{ins};x)$, which models the likelihood of various potential responses. The final response is refined through beam search, an algorithm that systematically explores a range of possible sequences to identify the one that maximizes the overall probability.

\section{Experiment}
\subsection{Experimental Setup}
\textbf{\textit{Datasets}}~~~~We selected high-resource and low-resource datasets as follows:
YELP \cite{luca2016reviews} includes parallel sentences of positive and negative reviews.
GYAFC \cite{rao-tetreault-2018-dear} provides parallel sentences of formal and informal expressions within the domains of Entertainment\&Music and Family\&Relationships, respectively. 
Shakespeare \cite{zhu2023storytrans} contains the works of Shakespeare for the study of stylistic transfer between old and modern English.
Genshin \cite{xu2023specializing} is based on game roles and contains \textbf{six sub-datasets}: Xiangling, Hutao, Mona, Diluc, Venti, and Noelle.

\begin{table}[!htb]
\begin{center}
{\caption{Statistics of our datasets after pre-processing. The top three are considered\textbf{ high-resource} styles and the bottom two are considered \textbf{low-resource} styles.}\label{data}}
\begin{tabular}{lcccc}
\toprule
\textbf{Dataset} &  \textbf{Attributes} & \textbf{Train} & \textbf{Validation} & \textbf{Test} \\
\hline
 YELP    & {Positive} & {440K} & {10K} & {3,000}  \\
            & {Negative} & {440K} & {10K} & {3,000}  \\
GYAFC E\&M & {Formal} & {52,595} & {2,877} & {1,416}  \\ 
            & {Informal}  & {52,595} & {2,356} & {1,082}  \\ 
GYAFC F\&R & {Formal}  & {51,967} & {2,788} & {1,332}  \\ 
            & {Informal}  & {51,967} & {2,247} & {1,019}  \\ 
            \hline
Shakespeare & {Auther} & {2322} & {290}& {290} \\
 Genshin$_\textsc{(6 Roles)}$  & {Roles} & {{1653}} & {{220}}& {{220}} \\
\bottomrule
\end{tabular}
\end{center}
\end{table}

\textbf{\textit{Baselines}}~~~~We compare our method with two types of methods, covering four unsupervised methods and seven supervised methods. The following baselines:
 \textsc{\textbf{CP-G \& CP-B}} \cite{xu2020variational} use VAEs to fix decoding issues in unsupervised text learning by constraining the posterior means in the latent space.
 \textsc{\textbf{TextSETTR}} \cite{riley2021textsettr} constructs a style vector space encoding many style features through unlabeled training.
  \textsc{\textbf{BTTS}} \cite{xu2023specializing} strengthens the style vector space through contrastive learning to extract complex features. 
 \textsc{\textbf{B-GST}} \cite{sudhakar-etal-2019-transforming} utilizes a method of supervised learning by leveraging the Transformer's internal mechanism to remove stylistic attributes from the source sentence.
 \textsc{\textbf{Stroytrans}} \cite{zhu2023storytrans} utilizes discourse representations to capture source content and transfer it to target styles using adaptable style embeddings.
We also selected \textsc{\textbf{BART}}$_\textsc{Base}$, \textsc{\textbf{BART}}$_\textsc{Large}$ \cite{lewis2019bart}, \textsc{\textbf{GPT-2}}$_\textsc{Large}$ \cite{brown2020language}, \textsc{\textbf{T5}}$_\textsc{Base}$, and \textsc{\textbf{T5}}$_\textsc{Large}$ \cite{raffel2023exploring} based on \textbf{fine-tuning} as baselines.

\begin{table}[!htb]
\centering
\caption{To assess the TST capabilities of LLMs, we examine three renowned TST models with publicly accessible human evaluations (\textbf{ARAE} \cite{zhao2018adversarially}, \textbf{CAAE} \cite{shen2017style},  \textbf{DAR} \cite{li2018delete}). The numbers represent Spearman's rank correlation coefficients, with higher values indicating greater consistency with human assessments. $^\Diamond$ represents the data from  \cite{ostheimer2023text}.}
\begin{tabular}{lrrrr}
\toprule
\textbf{Method} & \textbf{ARAE}& \textbf{CAAE}& \textbf{DAR}& \textbf{All} \\
\midrule[0.5pt]
\textsc{\textbf{fastText}}$^\Diamond$ & {0.498}& {0.550}& {0.332} & {0.473}  \\ 
\textsc{\textbf{TextCNN}}$^\Diamond$ & {0.512}   & {0.525}& {0.331}& {0.458}\\ 
\textsc{\textbf{BERTScore}}$^\Diamond$ & {0.513}  & {0.559}& {0.408}& {0.497}\\ 
 \hline
 \textsc{\textbf{Falcon-7b}}$^\Diamond$ & {-0.027}  & {-0.219}& {-0.118}& {-0.131} \\
\textsc{\textbf{Llama2-7b}}$^\Diamond$ & {0.091}  & {-0.128}& {-0.064} & {-0.039}\\ 
\textsc{\textbf{InsGPT}}$^\Diamond$ & {0.618}  & {0.543}& {0.584}& {0.574} \\ 
\textsc{\textbf{CG4}} & \textbf{0.659}  & \textbf{0.612}& \textbf{0.623}& \textbf{0.631} \\

\bottomrule
\end{tabular}

\label{llm-eval}
\end{table}

\textbf{\textit{Parameter Settings}}~~~~ We use the \textsc{\textbf{BART}}$_\textsc{Large}$ as the generation backbone. To process a series of source prompts, we set the prompt length to 200 and the learning rate to $1 \times 10^{-3}$. During fine-tuning, the learning rate during training is set to $3 \times 10^{-5}$, and the batch size is set to 16. We perform fine-tuning using NVIDIA's RTX4090. Each experiment involves training for at least 50 epochs. To ensure a fair comparison, the learning rate for all baselines is set at $3 \times 10^{-5}$. The value of $\lambda$ is set to 0.5 based on the performance in the validation set. We also refrain from using techniques like label smoothing, warm-up learning rates, and length penalties in our model and the baselines.

\textbf{\textit{Evaluation Metrics}}~~~~
To estimate \textit{Style Accuracy (ACC)}, previous work used a pre-trained BERT classifier called BERTScore \cite{zhang2020bertscore}.  In our approach, we use prompts to guide \textsc{\textbf{ChatGPT-4}} \cite{openai2023gpt4} in the evaluation process, which we refer to as \textbf{\textit{CG4}}. Specifically, following the approach proposed by  \cite{jafaritazehjani-etal-2020-style}, we feed the model with supervised sentences and their manually assigned scores to guide its assessment of stylistic accuracy. Additionally, we require the model to justify its scores, as  \cite{ostheimer2023text} found that this enhances the precision of evaluations. 

We benchmarked above method against fastText \cite{joulin2016bag}, TextCNN \cite{kim2014convolutional}, and BERTScore \cite{zhang2020bertscore} in tabel~\ref{llm-eval}. Results show that CG4 aligns more closely with human evaluations across all TST models. We briefly introduce the evaluation method in the main text and provide experimental details in Appendix~\ref{LLM Evaluation}.

To assess the \textit{Content Consistency (CC)} in the target style, we adopted methods proposed by  \cite{pmlr-v119-xu20a} and  \cite{post2018clarity}, utilizing SacreBLEU to calculate self-BLEU between the output and input. Following  \cite{xu-etal-2018-unpaired}, we used "G" (geometric mean of Acc and content) as the metric for assessing the overall model quality. 

\begin{table*}[!htb]
\centering
\caption{Comparison of Full Dataset in TST. \underline{Underlined} data shows SETTP versus backbone. \textbf{Bold} indicates the best, and \textbf{\textit{italicized}} the second. $^\heartsuit$ denotes average metrics for sub-tasks, and expanded results on \textbf{thirteen} TST tasks are shown in the Appendix~\ref{Expanded}.}\label{Style Transfer}
\begin{tabular}{lccc|ccc|ccc|ccc|ccc|c}
\toprule
\textbf{Dataset} & 
\multicolumn{3}{c}{YELP$^\heartsuit$}&
\multicolumn{3}{|c}{GYAFC E\&M$^\heartsuit$}&
\multicolumn{3}{|c}{GYAFC F\&R$^\heartsuit$} &
\multicolumn{3}{|c}{Shakespeare} &
\multicolumn{3}{|c|}{Genshin$^\heartsuit$}\\

\cmidrule(r){1-1}\cmidrule(r){2-4}\cmidrule(r){5-7}\cmidrule(r){8-10}\cmidrule(r){11-13}\cmidrule(r){14-17}
\textbf{Model} & \textbf{CC} & \textbf{ACC}& \textbf{G}& \textbf{CC} & \textbf{ACC}& \textbf{G}& \textbf{CC} & \textbf{ACC}& \textbf{G}& \textbf{CC} & \textbf{ACC}& \textbf{G}& \textbf{CC} & \textbf{ACC}& \textbf{G}& \textbf{avg.G} \\
\midrule[0.5pt]
      \textsc{\textbf{CP-G}}& 35.7  & 51.1  & 42.7  & 34.1  & 68.5  & 48.3  & 37.0  & 71.8  & 51.5  & - & - & - & - & - & - & \cellcolor{lightgray}47.5 \\ 
      \textsc{\textbf{CP-B}} & 36.5  & 51.5  & 43.4  & 40.1  & 72.5  & 53.9  & 42.8  & 76.1  & 57.1  & - & - & - & - & - & -  &\cellcolor{lightgray}51.4\\ 
      \textsc{\textbf{TEXTSETTR}} & 44.9  & 54.5  & 49.5  & 47.2  & 75.6  & 59.7  & 51.7  & 77.5  & 63.3  & 13.8  & 78.6  & 32.9  & 32.5  & 52.6  & 41.3  &\cellcolor{lightgray}49.4 \\ 
       \textsc{\textbf{BTTS}} & \textbf{54.7}  & 53.7  & 54.2  & 52.9  & 75.9  & 63.4  & 53.4  & 79.6  & 65.2  & \textbf{17.4}  & 74.6  & 36.0  & 47.6  & \textit{\textbf{62.7}}  & \textit{\textbf{54.6}}  &\cellcolor{lightgray}54.7 \\ 
       \hline
      \textsc{\textbf{BART}}$_\textsc{Base}$ & 42.1  & 53.7  & 47.5  & 32.5  & 67.2  & 46.7  & 34.6  & 62.4  & 46.5  & 10.9  & 71.4  & 27.9  & 36.5  & 47.6  & 41.7   &\cellcolor{lightgray}42.1\\ 
      \textsc{\textbf{BART}}$_\textsc{Large}$ & \underline{48.9}  & \underline{55.2}  & \underline{52.0}  & \underline{50.6}  & \underline{70.4}  & \underline{59.7}  & \underline{56.8}  & \underline{\textit{\textbf{82.6}}}  & \underline{68.5}  & \underline{12.7}  & \underline{72.6}  & \underline{30.4}  & \underline{48.7}  & \underline{56.3}  & \underline{52.4}  & \cellcolor{lightgray}51.7\\ 
      \textsc{\textbf{GPT-2}}$_\textsc{Large}$  & 32.1  & 48.6  & 39.5  & 30.7  & 56.2  & 41.5  & 31.7  & 52.7  & 40.9  & 14.6  & 67.4  & 31.4  & 31.5  & 41.2  & 36.0  &\cellcolor{lightgray}37.9 \\ 
      \textsc{\textbf{T5}}$_\textsc{Base}$ & 39.9  & 53.6  & 46.2  & 37.2  & 62.9  & 48.4  & 30.0  & 60.2  & 42.5  & 11.7  & 68.7  & 28.4  & 36.1  & 44.1  & 39.9   &\cellcolor{lightgray}41.1\\ 
      \textsc{\textbf{T5}}$_\textsc{Large}$ & 52.0  & 56.7  & 54.4  & 45.7  & 71.5  & 57.2  & 57.3  & 70.7  & 63.6  & 12.8  & 73.4  & 30.7  & \textit\textbf{{53.1}}  & 55.6  & 54.3 &\cellcolor{lightgray}52.0  \\ 
      \textsc{\textbf{B-GST}} & 52.1  & \textit{\textbf{57.8}}  & \textit{\textbf{54.9}}  & \textit{\textbf{53.4}}  & \textbf{80.0}  & \textit{\textbf{65.4}}  & \textit{\textbf{58.6}}  & 81.2  & \textit{\textbf{69.0}}  & \textit{\textbf{16.7}}  & 82.5  & \textit{\textbf{37.1}}  & 30.7  & 60.1  & 43.0   &\cellcolor{lightgray}\textit{53.8}\\ 
      \textsc{\textbf{Stroytrans}} & 50.7  & 53.4  & 52.0  & 51.9  & 57.2  & 54.5  & 52.5  & 49.9  & 51.2  & 12.7  & \textit{\textbf{83.4}}  & 32.5  & 52.1  & 46.7  & 49.3&   \cellcolor{lightgray}47.9\\ 
      \hline
      \textsc{\textbf{OURS}}$_\textsc{Base}$ & 42.7  & 59.7  & 50.5  & 49.4  & 61.7  & 55.2  & 51.5  & 66.0  & 58.3  & 15.6  & 81.3  & 35.6  & 40.3  & 46.1  & 43.1  &\cellcolor{lightgray}48.6 \\ 
      \textsc{\textbf{OURS}}$_\textsc{Large}$  & \underline{\textit{\textbf{54.5}}} & \underline{\textbf{62.3}}  & \underline{\textbf{58.3}}  & \underline{\textbf{54.7}}  & \underline{\textit{\textbf{79.2}}}  & \underline{\textbf{65.8}}  & \underline{\textbf{58.9}}  & \underline{\textbf{83.9}}  & \underline{\textbf{70.3}}  & \underline{16.5}  & \underline{\textbf{85.7}}  & \underline{\textbf{37.6}}  & \underline{\textbf{55.1}}  & \underline{\textbf{64.5}}  & \underline{\textbf{59.6}}   &\cellcolor{lightgray}\textbf{58.3}\\ 
\bottomrule
\end{tabular}

\end{table*}

\subsection{Full Dataset Setting}
\label{Full}
Table~\ref{Style Transfer} demonstrates SETTP's capability for style transfer through cross-style knowledge, employing all available training data from the target task. In our experiments, prompts extracted from the source prompt pool are transferred to the target style. For instance, we extract transferable representations as prompts from source domains YELP, GYAFC E\&M, GYAFC F\&R, and Genshin for training on the low-resource Shakespeare dataset.

\textbf{Results on High-resource Style Transfer}~~~~
 SETTP has achieved the highest G-score across these three high-resource datasets, with G-scores of 58.3, 65.8, and 70.3, respectively, marking improvements of 6.17\%, 0.61\%, and 1.88\% over the previous best methods. 
Surprisingly, the result demonstrates that transferable knowledge is also effective for well-resourced tasks.

\textbf{Results on Low-resource Style Transfer}~~~~In the context of low resources, the performance of models has significantly declined, which seems to be related to both the complex nature and quantity of the datasets. All methods exhibited low CC for the Shakespeare, while ACC was generally high. This may be due to using Early Modern English syntax, which challenges models. PLMs are primarily exposed to modern English during pre-training. On the Genshin, SETTP demonstrated a clear advantage over most baseline models. Significant improvements in more complex transfer tasks can be attributed to effective dual-level prompt learning. 

\textbf{Summary of Main Results}~~~~Overall, SETTP has achieved an 8.36\% improvement over the previous SOTA(average G-score: 53.8), reaching an average G-score of 58.3 across all different datasets. In high-resource TST tasks, backbone {\textsc{\textbf{BART}}$_\textsc{Large}$ achieved an average G-score of 60.0, while STEEP scored an average of 64.8, marking an 8.00\% increase. In low-resource TST, the average G-score for \textsc{\textbf{BART}}$_\textsc{Large}$ was 41.4, with STEEP reaching 48.6, an increase of 17.39\%. Hence, our initial findings indicate the need to incorporate external knowledge increases as dataset size decreases.

\begin{table*}[!htb]
\centering
  \caption{Comparison of the methods under the different numbers of training data.
  }\label{wo and few}
\begin{tabular}{l ccc|ccc}
\toprule
	\textbf{Dataset} & \multicolumn{3}{c}{Genshin$^\heartsuit$  \textbf{(CC/ACC/G)}} & \multicolumn{3}{|c}{GYAFC E\&M$^\heartsuit$  \textbf{(CC/ACC/G)}}  \\
\cmidrule(r){1-1}\cmidrule(r){2-7}
	\textbf{\# Instances} & \multicolumn{1}{c}{1\%} & \multicolumn{1}{c}{2\%} & \multicolumn{1}{c}{5\%} & \multicolumn{1}{|c}{1\%} & \multicolumn{1}{c}{2\%} & \multicolumn{1}{c}{5\%}  \\
\midrule[0.5pt]
	\textsc{\textbf{TextSETTR}} & \textbf{50.7}/29.8/38.9 & 45.3/32.7/38.5 & 43.5/36.7/40.0 & \textbf{50.4}/40.2/45.0 & 47.6/55.4/51.4 & 48.6/\textit{64.2}/55.9\\
	\textbf{\textsc{BTTS}} & 47.4/26.5/35.4 & 48.6/34.5/40.9 & 44.9/32.2/38.0 & 48.2/41.7/44.8 & 48.5/\textit{59.5}/\textit{53.7} & 49.1/63.8/56.0\\
	\textsc{\textbf{BART}}$_\textsc{Base}$  & 41.5/28.9/34.6 & 42.1/35.2/38.5 & 40.7/36.9/38.8 & 40.5/38.6/39.5 & 39.1/43.5/41.2 & 35.6/48.9/41.7\\
	\textsc{\textbf{BART}}$_\textsc{Large}$ &\textit{50.1}/32.6/\textit{40.4} & \textit{49.3}/37.5/\textit{43.0} & \textit{46.0}/39.8/\textit{42.8} & \textit{50.3}/43.7/\textit{46.9} & \textit{50.4}/53.5/51.9 & \textit{51.2}/62.8/\textit{56.7}\\
 	\textsc{\textbf{B-GST}} & 36.8/30.2/33.3 & 39.2/36.9/38.0 & 32.7/\textit{46.4}/39.0 & 42.8/\textit{44.2}/43.5 & 47.5/58.2/52.6 & 49.8/63.1/56.1\\
\cline{1-7}
	\textsc{\textbf{OURS}}$_\textsc{Base}$ & 40.5/\textit{35.7}/38.0 & 42.7/\textit{42.7}/42.7 & 41.5/43.2/42.3 & 42.7/40.6/41.6 & 40.6/44.2/42.4 & 43.6/50.6/47.0 \\
        \textsc{\textbf{OURS}}$_\textsc{Large}$ & 49.8/\textbf{42.6}/\textbf{46.1} & \textbf{50.3}/\textbf{49.6}/\textbf{49.9} & \textbf{48.6}/\textbf{53.7}/\textbf{51.1} & 49.5/\textbf{51.8}/\textbf{50.}6 & \textbf{51.0}/\textbf{65.9}/\textbf{58.0} & \textbf{53.7}/\textbf{74.9}/\textbf{63.4} \\
\bottomrule
\end{tabular}
\end{table*}

\subsection{Few-Shot Setting}
To further simulate extreme low-resource scenarios in actual environments, experiments are conducted using varying amounts of target domain data,  n\% of the full training instances. We randomly sample each size five times and take the average as the final result. To ensure fairness with other baselines, the data selection for the target domain should be limited to a randomly extracted range, which will serve as the basis for generating target style-level prompts.



\textbf{Results on Few Instances}~~~~Table~\ref{wo and few} indicates that SETTP achieves the best result among all the methods under limited training data.
Compared to the baseline model, SETTP has significantly improved few-shot learning capabilities. In the Genshin, when using 5\% instances, our performance (average G-score: 51.1) exceeded the best baseline (average G-score: 42.8) by 16.24\%.
In the high-resource dataset {GYAFC E\&M, we used about 250 parallel data to achieve a similar effect (G-score: 63.4) as the baselines did with over 50,000 (G-score: 65.4). 
As shown in Figure~\ref{few-fig}, SETTP is capable of effective learning when the data is extremely limited. However, as the quantity of data increases, this advantage gradually diminishes, which is consistent with the findings in Section~\ref{Full}.


\begin{table}[!htb]
\begin{center}
{\caption{Human Evaluation Metrics.}\label{human}}
\begin{tabular}{lccc|c}
\toprule
\textbf{Method} &  \textbf{Style} & \textbf{Content} & \textbf{Fluency} & avg \\
\hline
\textsc{\textbf{TEXTSETTR}}     & {4.02} & {4.14} & {\textbf{2.46}} & \cellcolor{lightgray}3.54 \\
\textsc{\textbf{BTTS}}          & {2.56} & {2.45} & {2.52}  & \cellcolor{lightgray}2.52\\ 
\textsc{\textbf{B-GST}}         & {1.96} & {3.46} & {3.52}  &\cellcolor{lightgray}2.98\\ 
\textsc{\textbf{Stroytrans}}    & {4.80} & {2.27} & {3.78} &\cellcolor{lightgray}3.62\\
 \textsc{\textbf{OURS}}$_\textsc{Large}$   & \textbf{1.66} & \textbf{2.28} & {2.82} &\cellcolor{lightgray}\textbf{2.25}\\
\bottomrule
\end{tabular}
\end{center}
\end{table}

\textbf{Human Evaluation}~~~~
To supplement automatic metrics, we perform human evaluations, sampling 50 instances from each dataset (YELP, GYAFC E\&M, GYAFC F\&R,  Shakespeare, Genshin). Fifty participants are enlisted to assess each model on three criteria: (1) strength of style transfer, (2) semantic integrity, and (3) sentence fluidity. Based on performance, models are ranked from best 1 to worst 5. Table~\ref{human} presents the average rankings of the five models, as determined by participant feedback. In the assessment of Content and Fluency, participants seem to struggle with making a choice; however, SETTP significantly leads in terms of Style Accuracy.

\section{Analysis}
\subsection{Ablation Study} 
We conduct an ablation study in Table~\ref{Ablation} to investigate the importance of the components in SETTP. "w/o \textit{AAR}" denotes ablating Adaptive Attentional Retrieval, i.e., Equations 4 and 5, and directly using target low-resource prompt $P_T$. "w/o \textit{Style}" denotes that we trained $f_{\theta_S}(\cdot)$ directly without any $P_T$."w/o \textit{DT}" indicates ablating Domain Temperature, i.e., changing $s_n$ in Equation 5 to a fixed value of $1/N$. w/o \textit{PI}" signifies ablating Prompt-tunable Inference, i.e., random prompts are used during inference. "w/o \textit{Cluster}" denotes ablating the clustering process, reducing the coupling of content and style.

\textbf{Effect of AAR and Style-level Prompt}~~~~ 
From Table~\ref{Ablation} (row 1 and row 2), using style-level prompts without interpolation and not using style-level prompts significantly diminish performance, particularly in ACC. The former demonstrates the effectiveness of knowledge transfer\underline{\textbf{(Challenge 1)}} via AAR, while the latter shows that the \textbf{\textit{Style-level Prompts}} aid in initializing training.

\textbf{Effect of PI and Instance-level Prompt}~~~~
From Table~\ref{Ablation} (row 4 and row 5), using random prompts or using prompts that do not cluster based on semantics both lead to a decline in content consistency, especially the latter, which aligns with our hypothesis that \textbf{\textit{Instance-level Prompts}} can reduce semantic bias \underline{\textbf{(Challenge 2)}}. We found that clusters of semantically similar data generated through clustering can reduce the impact of irrelevant content.

\textbf{Effect of Domain Temperature}~~~~
Eliminating \textit{DT} did not result in significant performance degradation, indicating that the improvement brought by Domain Temperature is valuable but limited.

\begin{table*}[!htb]
\centering
  \caption{The ablation study of the methods under the different numbers of training data. \textbf{Bold} fonts indicate the biggest performance drop after ablation and \textbf{\textit{italicized}} the second. 
 }\label{Ablation} 
\begin{tabular}{l ccc|ccc}
\toprule
	\textbf{Dataset} & \multicolumn{3}{c}{Genshin$^\heartsuit$  \textbf{(CC/ACC/G)}} & \multicolumn{3}{|c}{GYAFC E\&M$^\heartsuit$  \textbf{(CC/ACC/G)}}  \\
\cmidrule(r){1-1}\cmidrule(r){2-7}
	\textbf{\# Instances} & \multicolumn{1}{c}{1\%} & \multicolumn{1}{c}{2\%} & \multicolumn{1}{c}{5\%} & \multicolumn{1}{|c}{1\%} & \multicolumn{1}{c}{2\%} & \multicolumn{1}{c}{5\%}  \\
\midrule[0.5pt]
        \textsc{\textbf{OURS}$_\textsc{Large}$  w/o  \textit{AAR}} & 51.7/\textbf{\textit{33.6}}/\textbf{\textit{41.7}} & 51.3/\textbf{\textit{40.2}}/\textbf{\textit{45.4}} & 50.3/\textit{\textbf{48.6}}/\textbf{\textit{49.4}} & 49.6/\textbf{\textit{45.5}}/\textbf{\textit{47.5}} & 52.2/\textbf{\textit{56.7}}/\textbf{\textit{54.4}} & 51.3/\textbf{\textit{64.8}}/\textbf{\textit{57.7}}\\
        
        \textsc{\textbf{OURS}$_\textsc{Large}$  w/o  \textit{Style}} &51.3/\textbf{33.2}/\textbf{41.3}  & 51.2/\textbf{38.0}/\textbf{44.1}  & 51.0/\textbf{46.1}/\textbf{48.5}  & 49.7/\textbf{45.1}/\textbf{47.3}  & 52.1/\textbf{55.6}/\textbf{53.8}  & 50.9/\textbf{64.3}/\textbf{57.2}\\ 
        
        \textsc{\textbf{OURS}$_\textsc{Large}$  w/o  \textit{DT}} & 51.2/40.6/45.6 & \textbf{51.1}/44.5/47.7 & 49.8/50.8/50.3 & 50.1/50.2/50.1 & 51.5/59.6/55.4 & 53.4/66.9/59.8 \\ 
        
        \textsc{\textbf{OURS}$_\textsc{Large}$  w/o  \textit{PI}} & \textbf{50.3}/41.2/45.5 & 52.1/46.8/49.4 & \textbf{\textit{48.7}}/52.2/50.4 & \textbf{\textit{47.6}}/47.6/47.6 & \textbf{48.9}/63.2/55.6 & \textbf{\textit{50.2}}/71.5/59.9\\
        
        \textsc{\textbf{OURS}$_\textsc{Large}$  w/o  \textit{Cluster}} & \textit{\textbf{50.4}}/40.5/45.2  & 51.3/45.3/48.2  & \textbf{48.5}/51.7/50.1  &\textbf{47.3}/47.5/47.4  & \textbf{\textit{49.1}}/63.7/55.9  &\textbf{49.7}/70.5/59.2  \\ 
        
\cline{1-7}
        \textsc{\textbf{OURS}}$_\textsc{Large}$ & \underline{49.8/42.6/46.1} & \underline{50.3/49.6/49.9} & \underline{48.6/53.7/51.1} & \underline{49.5/51.8/50.6} & \underline{51.0/65.9/58.0} & \underline{53.7/74.9/63.4} \\
        
\bottomrule
\end{tabular}
\end{table*}

\subsection{Visualization}
\textbf{Stylistic Feature}~~~~
We visualize the stylistic features of source texts, golden target texts, and transformed texts by SETTP via 2D UMAP \cite{mcinnes2020umap} in Figure~\ref{umap}. Due to space limitations, we showcase the results on a low-resource dataset from Genshin. We selected 12 features for visualization dimensions: these include three stylistic features, i.e., quantities of punctuation, the number of sentences, and the number of words \citep{zhu2023storytrans}; and nine less-correlated vertical style types\citep{kang2021style}, i.e., Humorous, Polite, Formal, Romantic, Gender, Dominance, Exciting, Sadness, and Offense. The first set of features are determined statistically, while the latter are evaluated using a BERT classifier  \citep{kang2021style}.

 The results clearly illustrate that SETTP’s transformed texts (in {\color{lightblue}BLUE}) are distinctly different from the original text style (in {\color{lightred}RED}) and closely align with the supervised target results (in {\color{lightyellow}GOLDEN}).

\begin{figure}[htb]

  \begin{subfigure}{.15\textwidth}
    \centering
    \includegraphics[width=\linewidth]{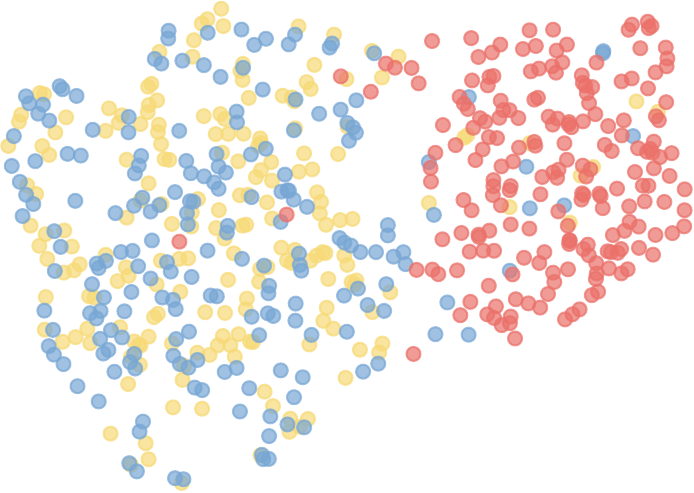}
    \caption{Hutao}
  \end{subfigure}
  \hfill
    \begin{subfigure}{.15\textwidth}
    \centering
    \includegraphics[width=\linewidth]{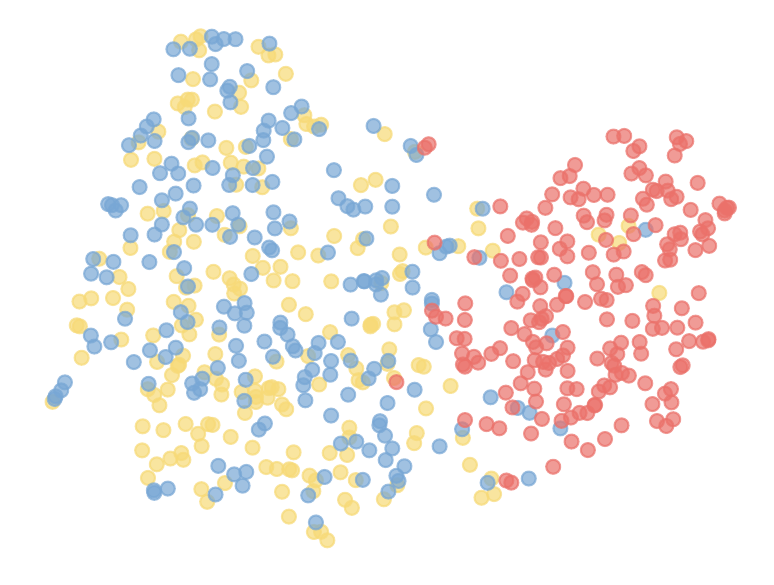}
    \caption{Venti}
  \end{subfigure}
    \hfill
  \begin{subfigure}{.15\textwidth}
    \centering
    \includegraphics[width=\linewidth]{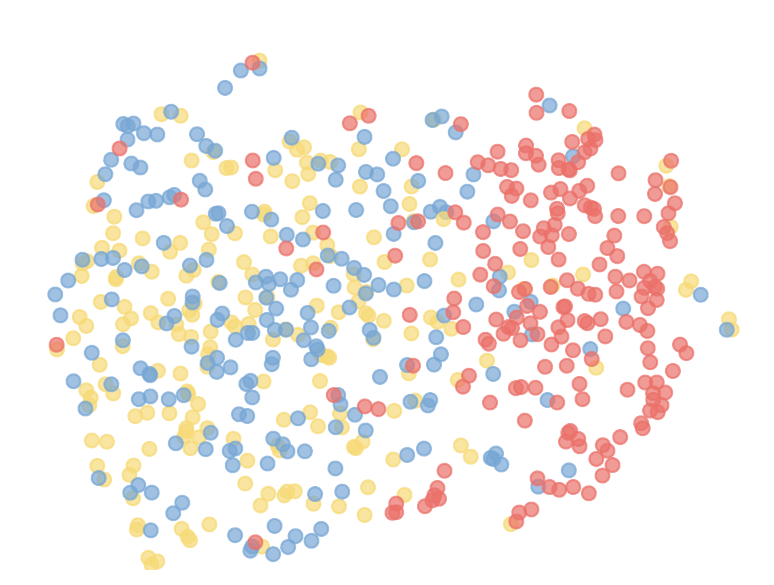}
    \caption{Diluc}
  \end{subfigure}
  \hfill
  
  \begin{subfigure}{.15\textwidth}
    \centering
    \includegraphics[width=\linewidth]{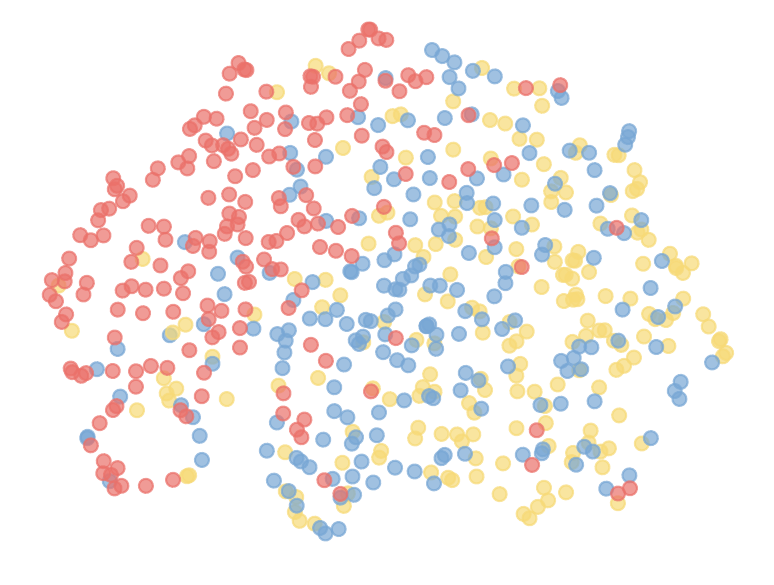}
        \caption{Noelle}
  \end{subfigure}
  \hfill
    \begin{subfigure}{.15\textwidth}
    \centering
    \includegraphics[width=\linewidth]{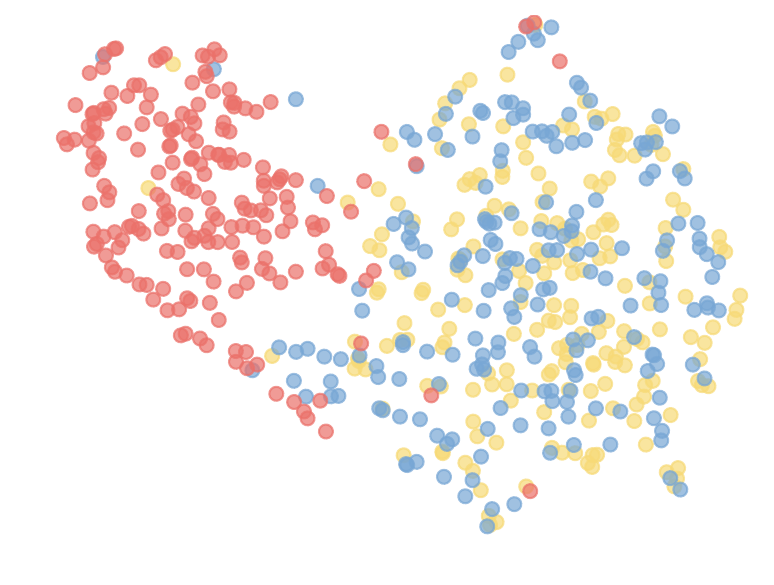}
        \caption{Mona}
  \end{subfigure}
  \hfill
  \begin{subfigure}{.15\textwidth}
    \centering
    \includegraphics[width=\linewidth]{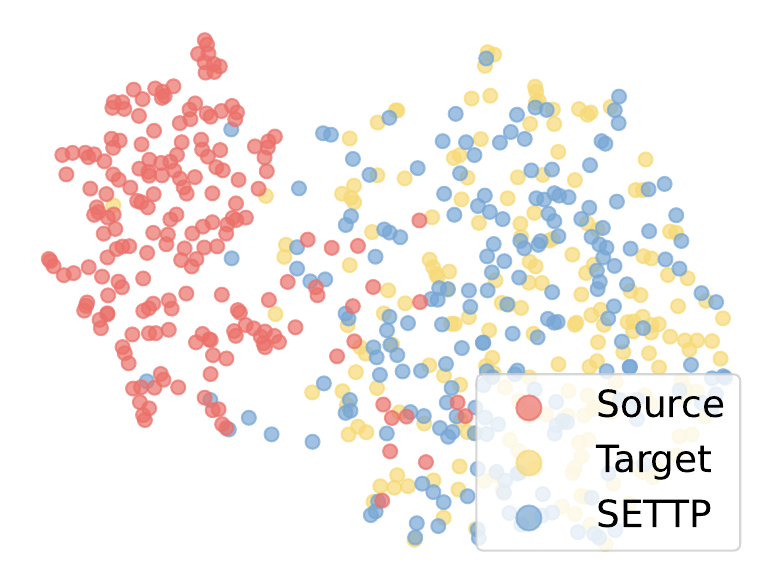}
        \caption{Xiangling}
  \end{subfigure}
  
\caption{2D-Visualization of Stylistic Feature. }\label{umap}
\end{figure}

\begin{figure}[!htb]
    \centering
    \includegraphics[width=0.3\textwidth]{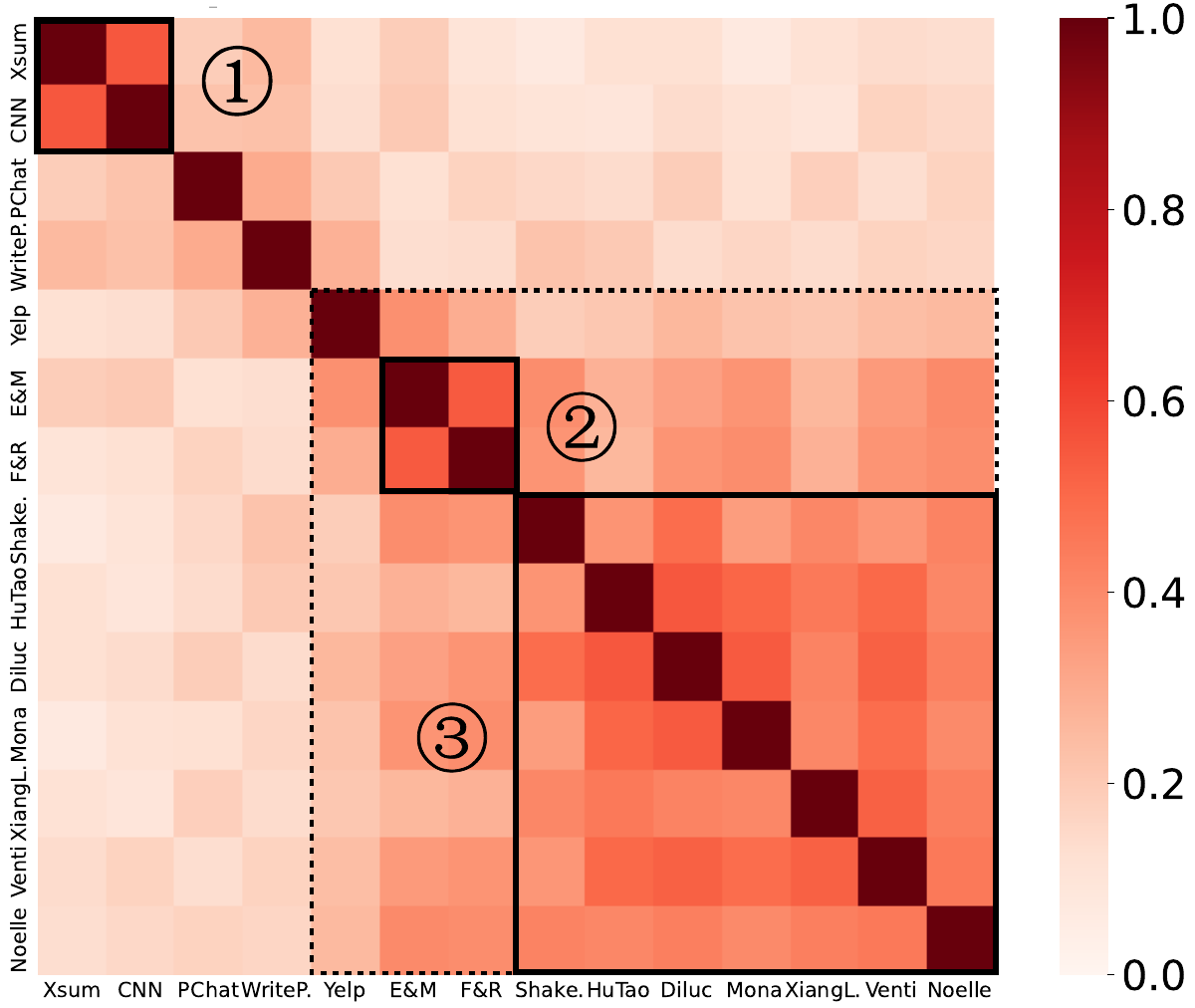}
    \caption{The confusion matrix for 14 datasets, where a deeper color indicates a higher similarity. \ding{172} corresponds to two summarization tasks, showcasing a high level of internal similarity; \ding{173} represents two style transfer tasks between formal and informal styles, indicating a high similarity; \ding{174} involves the Shakespeare and Genshin Impact datasets, also showing a high similarity. Observations from the dashed-line areas reveal that the similarity among style transfer tasks is generally higher than that of other tasks.}
\label{heatmap_prompt}
\end{figure}

\textbf{Task Similarity}~~~~
\label{Visualizationp}
Our approach is inspired by transferable prompt learning, leveraging multi-task prompts for transfer learning. Initial experiments~\cite{li2022learning} indicate that tasks with low relevance do not effectively transfer knowledge, whereas multi-style prompts exhibit higher similarity. 

Figure~\ref{heatmap_prompt} presents a clustered heatmap of the cosine similarity between multi-task and multi-style prompts. We further examined three additional tasks: story writing (WritingPrompts), summarization (Xsum, CNN/Daily Mail), and dialogue (PersonaChat). The results show higher similarities within the same tasks, for instance, between Xsum and CNN/Daily Mail; similarly, high similarities are observed in style transfers, especially among the six characters of GENSHIN. 
This observation confirms our conclusion: in our method, TST tasks can be mutually enhanced by learning and applying specific style cues to the target style. This suggests that transferable styles can be identified effectively through this method.



\textbf{Case Study}~~~~
We also provide some examples in Appendix~\ref{casestudy} to illustrate the effectiveness of SETTP.

\section{Comparison with advanced LLMs}
Fine-tuning LLMs often requires high-resource datasets to avoid overfitting. Closed-source LLMs (e.g., ChatGPT, Claude3\footnote{The specific versions of LLMs we used are gpt-3.5-turbo-1106, gpt-4-1106 and claude-3-sonnet-20240229.}) have demonstrated strong Zero-Shot capabilities. Still, they are constrained by high interface costs and potential privacy leaks during transmission. We compare commonly used LLMs with our method using the Genshin dataset. In addition to the evaluation metrics for TST, we also tested the average memory usage of LLMs on the edge side and the token cost of LLMs on the cloud side.

Table~\ref{LLMs} demonstrates that SETTP has achieved performance comparable to SFT \textsc{\textbf{Llama2-7b}} and \textsc{\textbf{Llama2-13b}},  with significantly lower memory consumption. Additionally, our method's results are closely similar to those of Zero-Shot approaches, particularly \textsc{\textbf{ChatGPT-3.5}}, which shows obvious weakness in the CC metrics. A notable gap exists when compared to the Few-Shot approaches. 
This divergence is expected due to the considerably higher operational costs (API expenses, parameter requirements) associated with Closed-source LLMs. Our cost-effective solution can partially replace the Zero-Shot approach, especially in environments with limited internet and computational resources. Experimental details are provided in Appendix~\ref{LLMs4TST}.

\begin{table}[htb]
\centering
\caption{SFT stands for Supervised Fine-Tuning model. Zero-shot and Few-Shot mean learning from examples contained in prompts. Notably, {$^\triangle$} represents possible \textbf{Self-bias} as our evaluation is based on \textsc{\textbf{ChatGPT-4}}. 110T (in \& out) means that each conversation input message and reply together consume 110 tokens.}\label{LLMs}
\begin{tabular}{llll|c}
\toprule
\textbf{Dataset} &  \multicolumn{4}{c}{Genshin}\\

\cmidrule(r){1-1}\cmidrule(r){2-5}
\textbf{Method} & \textbf{CC} & \textbf{ACC}& \textbf{G}& \textbf{Inference Cost}\\
\midrule[0.5pt]
      SFT \textsc{\textbf{Llama2-7b}}& 56.9  & 63.7$\downarrow$ & 60.2 & 14.3GB VRAM \\
      SFT \textsc{\textbf{Llama2-13b}}& 54.7$\downarrow$  & 69.5  & 61.7& 27.1GB VRAM\\ 
      \hline
      Zero-Shot \textsc{\textbf{ChatGPT-4}}{$^\triangle$}& 51.8$\downarrow$  & 70.1  & 60.3  & 110T (in \& out)\\
      Zero-Shot \textsc{\textbf{ChatGPT-3.5}}& 45.8$\downarrow$  & 68.3  & 55.9$\downarrow$ &   110T (in \& out)\\
      Zero-Shot \textsc{\textbf{Claude3}}& 53.6$\downarrow$  & 67.0  & 59.9 &107T (in \& out) \\
      Few-Shot \textsc{\textbf{ChatGPT-4}}{$^\triangle$}& 58.9  & \textbf{83.9}  & \textbf{70.3} & 160T (in \& out) \\
      Few-Shot \textsc{\textbf{ChatGPT-3.5}}& 53.8$\downarrow$  & 72.8  & 62.6  & 160T (in \& out)\\
      Few-Shot \textsc{\textbf{Claude3}}& \textbf{60.7}  & 78.6  & 69.1  & 160T (in \& out)\\
    \hline
      \textsc{\textbf{OURS}}$_\textsc{Large}$  & \underline{55.0}  & \underline{64.5}  & \underline{59.6}  & 3.2GB VRAM\\
    
\bottomrule
\end{tabular}
\end{table}



\section{Related Work}
\textbf{\textit{Text Style Transfer}} has been significantly propelled forward, leveraging the robust generalization capabilities of PLMs. Recent TST techniques are mainly categorized into supervised and unsupervised methods. Supervised methods rely on extensive annotated data, and models well-tuned with sufficient data often achieve exceptional performance \cite{luca2016reviews} \cite{sudhakar-etal-2019-transforming}. However, annotated data for specific styles tend to be scarce in natural settings, leading many studies to use unsupervised methods \cite{rao-tetreault-2018-dear} \cite{zhu2023storytrans}. These methods reduce the reliance on annotated data but still require a certain amount of data, and their performance typically falls short of supervised methods \cite{pmlr-v119-xu20a}.

 \textbf{\textit{Prompt}} is a method of controlling PLMs. Early prompt learning focused on manually crafted \textbf{\textit{hand-crafted prompts}} \cite{raffel2023exploring}, whose limitation lies in the rigidity of the prompt templates, unable to adapt flexibly to different tasks and requirements. Some studies have proposed methods for optimizing continuous prompts \cite{liu2023gpt}, achieving optimal results in both continuous and discrete spaces. 
 The \textbf{\textit{soft prompts}} are essentially adjustable vectors or embeddings that guide the model to generate text meeting specific requirements \cite{asai2022attempt} \cite{li2022learning}. 
 The recently popular \textbf{\textit{prompt engineering}} \cite{brown2020language} is an application in instruct learning where independent prompt engineering does not affect the adjustment of model parameters, focusing instead on leveraging the current capabilities of LLMs.

\section{Conclusion}
We present a framework named SETTP to extract style-level prompts from high-resource styles using a specialized attention mechanism. It also proposes instance-level prompts to reduce semantic bias. Experiments show that SETTP achieves comparable performance to SOTA methods with only 1/20th of the data volume. 

\begin{ack}
This research was supported by the National Natural Science Foundation of China (No.62076059) and the Science and Technology Joint Project of Liaoning province (2023JH2/101700367, ZX20240193). Osmar Zaiane acknowledges the funding from NSERC and the Canada CIFAR AI Chairs Program.
\end{ack}

\bibliography{ecai}

\newpage
\appendix
\section{Domain Temperature}

In neural machine translation, researchers have introduced MultiUAT to address the imbalance between high-resource and low-resource domains in machine learning aimed at maintaining balance across different domains. This approach involves a dynamic adjustment strategy where the model's use of training data is adjusted based on the uncertainty associated with a small set of highly credible clean data. This method uses uncertainty metrics to regulate the data from each domain during training, allowing the model to effectively learn from high-resource and low-resource domains without bias towards any domain. In Section~\ref{DT}, we have further applied this approach to the field of style transfer, simplifying it as a means to constrain low-resource styles.

Precisely, we measure the cosine similarity between each content embedding $\bm{t}_S^l$ in the source prompt pool and $\bm{t}_T^j$ in the target prompt pool, as well as their respective prompt words $E_l^{ins}$ and $E_j^{ins}$. A similarity that exceeds certain thresholds ($\theta_t$ and $\theta_e$) indicates a high reference value. Through this exhaustive statistical method, we have learned a parameter for each source domain, referred to as "domain temperature," which helps us balance the availability and quality differences between domains, ensuring that low-resource settings do not overly impact the overall model performance.

\begin{algorithm} [!htb]
   \fontsize{8}{9}\selectfont
     \caption{Domain Temperature} 
      \KwIn{Content threshold $\theta_t$, Embedding threshold $\theta_e$, $\mathcal{P}^{style}$, $\mathcal{P}^{ins}$ }
      \KwOut{$\bm{w}=\{w_1,w_2,w_3...w_N\}$} 
        \While{$j < (n+1) $}
        { 
             Initialize $ctSim$, $ebdSim$, and $Sim$\;
            \For{$D_k \in \mathcal{P} $} 
            {  
             \For{$\bm{t}_l \in D_k $}{

                \If{$\|\bm{t}_l,\bm{t}_j\| \ge ctSim)\wedge \|E_l^\text{ins},E_j^\text{ins}\| \ge ebdSim$} 
                { 
                   $contSim \leftarrow \|\bm{t}_l,\bm{t}_j\|$\;
                   $ebdSim \leftarrow \|E_l^\text{ins},E_j^\text{ins}\|$\; 
                } 
              }
              \If{$contSim \ge \theta_t\wedge ebdSim \ge \theta_e$} 
                { 
                   $Sim=k$\;
                }
            } 
            \If{$Sim \neq null$} 
                { 
                   $w_{Sim} \leftarrow w_{Sim}+1$\;
                }
            $j \leftarrow j+1$
        } 
        
        return softmax$(\bm{w})$\; 
        \label{al}
\end{algorithm}

\section{LLM Evaluation for Style Accuracy}
\label{LLM Evaluation}
To estimate the \textit{\textbf{Style Accuracy (ACC)}}, We use prompts to guide LLMs for evaluation. Specifically, we followed the ideas of  \cite{jafaritazehjani-etal-2020-style}. And designed a series of prompts to guide the evaluation of LLMs. We tested the performance of large-scale models using consistency judgments from human evaluators. Furthermore, we discussed and proposed viable solutions to the biases that may arise in automatic evaluation, such as sequence bias or self-generation bias, to ensure evaluation accuracy. 

\subsection{Comparison of Classifier Methods }
We first conduct a comparative experiment using the classifier method. We examine the performance differences between small-scale Pretrained Language Models like BERT, RoBERTa, MacBERT, and LLMs. For small-scale PLMs, we extensively fine-tune them using datasets such as GYAFC, Yelp, Shakespeare, and Genshin. In contrast, LLMs rely on the knowledge embedded within their pre-training corpora. The experimental results are presented in Table~\ref{bert-gpt}.

This study presents a comparative analysis of BERT and ChatGPT models, focusing on their classification accuracy across high-resource and low-resource datasets. The results indicate that both models exhibit high accuracy levels when dealing with high-resource datasets. However, there is a notable decrease in accuracy for more low-resource styles for small-scale pre-trained language models. In contrast, the evaluation performance of LLMs like ChatGPT remains consistently high, demonstrating their robustness in handling low-resource datasets.

\begin{table}
\centering
\caption{The table describes the classification accuracy values that serve as the basis for classifiers comparison.}
\begin{tabular}{lcc|cc}
\toprule
\textbf{} & \multicolumn{2}{c}{\textbf{High-resource}} & \multicolumn{2}{c}{\textbf{Low-resource}}\\
\hline
\textbf{Model} & GYAFC & YELP & Genshin& \textsc{\textbf{Shakesp.}}\\
\hline
\textsc{\textbf{BERT}} & {89.71} & {91.33}& {32.61}& {42.71}\\
\textsc{\textbf{RoBERTa}} & {92.63} & {93.31} & {39.52}& {52.69}  \\
\textsc{\textbf{MacBERT}} & {88.92} & {92.84} & {37.18} & {55.32} \\
\hline
\textsc{\textbf{ChatGPT-4}}& \textbf{99.21} & \textbf{98.69} & \textbf{76.22} & \textbf{82.11}\\ 
\textsc{\textbf{ChatGPT-3.5}} & {98.28} & {97.33} & {69.35} & {80.35} \\ 
\bottomrule
\end{tabular}

\label{bert-gpt}
\end{table}

\subsection{Comparison of Rating Methods}
\subsubsection{Methods Description}
We have designed several evaluation strategies based on ChatGPT or similar LLM, and different strategies use different prompts. The prompts below can be found in Table ~\ref{prompt_eval}.

{\bf No-Reference Explicit Rating(NER)}
The core of this method relies on LLM's huge training corpus and uses prior knowledge to make judgments. By using given input texts, we guide ChatGPT to directly generate ratings, independently assessing the absolute quality of each text in specific aspects or overall. It is worth noting that if the model gives detailed reasons, the evaluation results will be more consistent with humans, so we require the model to give a specific analysis. 

{\bf Reference-Based Explicit Rating(RER)}
Using given input texts, we conversationally guide ChatGPT in generating ratings directly. Specifically, we pre-design the answers of the LLM and use them as input texts. As known, LLMs provide responses based on previous conversation history. By providing preset prior answers, we can effectively guide the model in evaluation. 

{\bf Paired Comparison Rating(PCR)}
This method focuses on assessing the relative quality of texts. We employed a cross-blind casting strategy to overcome biases such as order and self-generation (where models tend to prefer their own generated answers). 

Specifically, we conducted blind casting by randomly exchanging the order of answers generated by various models with the 'gold standard' answers. The voting criteria are categorized into (strong win, weak win, tie, weak loss, strong loss), corresponding to scores of (50, 40, 30, 20, 10). Ultimately, the average score of each stage is used as the measurement standard.

The final score for each passage is calculated as follows:
\begin{equation}
Score = \sum_{d=1}^{n} (Score_1 + Score_2)
\end{equation}
Here, \( n \) represents the number of models participating in the voting process. In this study, we have selected four different LLMs for voting, namely \textsc{\textbf{ChatGPT-4}} and \textsc{\textbf{LLaMA2-13B}}. This multi-model voting approach is aimed at ensuring objectivity and accuracy in scoring.

\subsubsection{Benchmark for human evaluation}
To assess the evaluation capabilities of law masters in TST, we considered three renowned TST models with publicly available human evaluation results.

\textsc{\textbf{Cross-Aligned Autoencoder (CAAE)}}  \cite{shen2017style}: An autoencoder model for style transfer.

\textsc{\textbf{Adversarially Regularized Autoencoder (ARAE)}}  \cite{zhao2018adversarially}: An autoencoder optimized through adversarial regularization.

\textsc{\textbf{Delete-and-Retrieve approach (DAR)}}  \cite{li2018delete}: A model implements style transfer through deletion and retrieval strategies.

We had LLMs evaluate the outputs from these models and calculate their correlation with human assessments to demonstrate their performance relative to existing automated metrics. For each model, we assessed the human-annotated sentences provided for the Yelp dataset, which includes an equal number of positive and negative examples from the test set (244 each), totaling 732 sentences.

\subsubsection{Correlations of LLM Evaluations with Human Evaluations}
To assess the effectiveness of LLMs in TST evaluation, we compare the correlation between the assessments produced by LLMs and the corresponding human evaluations across various aspects. 

Table~\ref{app-hunman} presents a comparison of various methods. \textsc{\textbf{CG4-NER}} represents the evaluation tests conducted on the \textsc{\textbf{ChatGPT-4}} platform using the No-Reference Explicit Rating (NER) strategy; \textsc{\textbf{CG3.5-RER}} refers to evaluations performed on ChatGPT-3.5-Turbo using the Reference-Based Explicit Rating (RER) strategy. Additionally, \textsc{\textbf{PCR}} denotes the mixed results obtained using the Paired Comparison Rating (PCR) strategy on \textsc{\textbf{ChatGPT-4}} and LLaMA-2 13B. It is important to note that the symbol $^\Diamond$ indicates that we directly used the data from  \cite{ostheimer2023text}.

The experiment result shows that the \textsc{\textbf{CG4-RER}} demonstrated high consistency with human performance, therefore we chose \textsc{\textbf{CG4-NER}} to represent \textsc{\textbf{CG4}} in the main text.

\begin{table}[!htb]
\centering
\caption{To assess the TST capabilities of LLMs, we examine three renowned TST models with publicly accessible human evaluations (\textbf{ARAE} \cite{zhao2018adversarially}, \textbf{CAAE} \cite{shen2017style},  \textbf{DAR} \cite{li2018delete}). The numbers represent Spearman's rank correlation coefficients, with higher values indicating greater consistency with human assessments. }
\begin{tabular}{lrrrr}
\toprule
\textbf{Method} & \textbf{ARAE}& \textbf{CAAE}& \textbf{DAR}& \textbf{All} \\
\midrule[0.5pt]
\textsc{\textbf{fastText}}$^\Diamond$ & {0.498}& {0.550}& {0.332} & {0.473}  \\ 
\textsc{\textbf{TextCNN}}$^\Diamond$ & {0.512}   & {0.525}& {0.331}& {0.458}\\ 
\textsc{\textbf{BERTScore}}$^\Diamond$ & {0.513}  & {0.559}& {0.408}& {0.497}\\ 
 \hline
 \textsc{\textbf{Falcon-7b}}$^\Diamond$ & {-0.027}  & {-0.219}& {-0.118}& {-0.131} \\
\textsc{\textbf{Falcon-40b}}$^\Diamond$ &    {0.206}  & {0.389}& {0.313}& {0.307} \\
\textsc{\textbf{Llama2-7b}}$^\Diamond$ & {0.091}  & {-0.128}& {-0.064} & {-0.039}\\ 
\textsc{\textbf{Llama2-13b}}$^\Diamond$ &    {0.103}  & {0.018}& {0.106}& {0.067} \\
\textsc{\textbf{InsGPT}}$^\Diamond$ & {0.618}  & {0.543}& {0.584}& {0.574} \\ 
 \hline
\textsc{\textbf{CG3.5-NER}}& {0.275}  & {0.262}& {0.281}& {0.273}\\ 
\textsc{\textbf{CG3.5-RER}} & {0.356}  & {0.392}& {0.307}& {0.352}\\ 
\textsc{\textbf{CG4-NER}} & {0.315}  & {0.279}& {0.361}& {0.318}\\ 
\textsc{\textbf{CG4-RER}} & \textbf{0.659}  & \textbf{0.612}& \textbf{0.623}& \textbf{0.631}\\ 
\textsc{\textbf{PCR}} & {0.535}  & {0.512}& {0.512}& {0.519}\\ 

\bottomrule
\end{tabular}

\label{app-hunman}
\end{table}

\section{Comparison with Advanced LLMs}
\label{LLMs4TST}
\subsection{SFT Methods Setting}
To demonstrate our model's strong generalization capability under low-resource conditions, we conducted a comparison with the renowned large language model, \textsc{\textbf{Llama2-7B}} and \textsc{\textbf{Llama2-13B}}. We employed a \textsc{\textbf{QLoRA}}-based supervised fine-tuning approach to ensure a fair comparison. 

Specifically, the parameter $lora_r$ determines the rank of the update matrix $BA$, which we set to a small rank of 1. When updating the parameter $w_0$, we modulate the impact of $BA$ using a scaling factor $\alpha$, which also acts as the learning rate, referred to as $lora_{\alpha}$. We also introduce $lora dropout$, a typical dropout strategy used for regularization. The number of training epochs is set to 3. Regarding the settings for $fp16$ and $bf16$, we choose to set both to $false$ since we do not use mixed precision training when employing \textsc{\textbf{QLoRA}}. The learning rate $r$ is set to $2 \times 10^{-4}$. All other settings are kept at their default values. We
perform tuning using 2 $\times$ A100 (80G). 

\subsection{Zero-Shot and Few-Shot Methods Setting}
The current closed-source LLMs demonstrate outstanding zero-shot and few-shot performance, but this depends on carefully crafted prompt engineering. In Table~\ref{prompt_gen}, we have designed a series of prompt schemes aimed at fully harnessing the potential of these large models, which include: \textsc{\textbf{ChatGPT-4}}, \textsc{\textbf{ChatGPT-3.5}}, and \textsc{\textbf{Claude3}}.

\subsection{Cost Comparison}
Table~\ref{LLMscompare} displays the performance comparison of the \textsc{\textbf{Llama2}} series fine-tuned with\textsc{\textbf{ Qlora}}, \textsc{\textbf{ChatGPT-4}},\textsc{\textbf{ ChatGPT-3.5}}, and \textsc{\textbf{CLAUDE3}} adjusted with prompting techniques, against our method. We conducted evaluations on two datasets, Genshin and GYAFC E\&M. In addition to comparing content consistency (CC) and style accuracy (ACC), we also analyzed Inference Cost and Inference Speed.

\begin{table*}[htb]
\centering
\caption{Comparison of advantages and disadvantages with LLMs. SFT stands for Supervised Fine-Tuning model. Zero-shot and Few-Shot means learning from examples contained in prompts. Notably, {$^\triangle$} represents possible \textbf{self-bias} as our evaluation is based on \textsc{\textbf{ChatGPT-4}}. 110T (in \& out) means that each conversation input message and reply together consume 110 tokens. $\downarrow$ means the score is lower than ours, and $^\star$ means results are subject to network fluctuations}\label{LLMscompare}
\begin{tabular}{llll|lll|ccr}
\toprule
\textbf{Dataset} &  \multicolumn{3}{c}{Genshin} &\multicolumn{3}{c}{GYAFC E\&M} &  \multicolumn{3}{c}{\textsc{\textbf{Average Cost}}}\\

\cmidrule(r){1-1}\cmidrule(r){2-10}
\textbf{Method} & \textbf{CC} & \textbf{ACC}& \textbf{G}&  \textbf{CC} & \textbf{ACC}& \textbf{G}& \textbf{Memory}& \textbf{Speed}& \textbf{Param}\\
\midrule[0.5pt]
      QLora-based SFT \textsc{\textbf{Llama2-7b}}& 56.9  & 63.7$\downarrow$ & 60.2 &57.8
&82.8&69.2& 14.3GB VRAM & 25.5T/S & 7B\\
      QLora-based SFT \textsc{\textbf{Llama2-13b}}& 54.7$\downarrow$  & 69.5  & 61.7&56.9&83.7&69.0& 27.1GB VRAM  & 17.2T/S & 13B\\ 
      \hline
      Zero-Shot Prompt 1 \textsc{\textbf{ChatGPT-4}}{$^\triangle$}& 51.8$\downarrow$  & 70.1  & 60.3  &51.0$\downarrow$&85.6&66.1& 47T (in \& out) & 5.4T/S$^\star$ & \\
      Zero-Shot Prompt 1 \textsc{\textbf{ChatGPT-3.5}}& 45.8$\downarrow$  & 68.3  & 55.9$\downarrow$ &53.2$\downarrow$&78.9$\downarrow$&64.8$\downarrow$& 47T (in \& out)& 8.7T/S$^\star$ &1750B\\
      Zero-Shot Prompt 1 \textsc{\textbf{Claude3}} & 53.6$\downarrow$  & 65.7  & 69.3 &55.7&83.5&68.2& 45T (in \& out) \\
          \hline
      Few-Shot Prompt 2 \textsc{\textbf{ChatGPT-4}}{$^\triangle$}& 48.2$\downarrow$  & 73.9  & 59.7 &48.7$\downarrow$&86.2&64.8& 120T (in \& out)& 5.4T/S$^\star$ \\
      Few-Shot Prompt 2 \textsc{\textbf{ChatGPT-3.5}}& 45.9$\downarrow$  & 68.2  & 55.9 &53.2$\downarrow$&83.3&66.5&  120T (in \& out)& 8.7T/S$^\star$ &1750B\\
      Few-Shot Prompt 2 \textsc{\textbf{Claude3}}& 50.2$\downarrow$  & 72.7  & 60.4  &51.7$\downarrow$&82.2&65.2$\downarrow$& 120T (in \& out)\\
        Few-Shot Prompt 3 \textsc{\textbf{ChatGPT-4}}{$^\triangle$}& 55.9  & 81.2  & 67.4 &58.7&85.9&71.0& 180T (in \& out) & 5.4T/S$^\star$\\
      Few-Shot Prompt 3 \textsc{\textbf{ChatGPT-3.5}}& 52.7$\downarrow$  & 70.3  & 60.9  &59.5&84.7&70.0& 180T (in \& out)& 8.7T/S$^\star$ &1750B\\
      Few-Shot Prompt 3 \textsc{\textbf{Claude3}}& 52.7$\downarrow$  & 76.6  & 63.5  &53.9$\downarrow$&86.8&68.4& 180T (in \& out)\\
    Few-Shot Prompt 4 \textsc{\textbf{ChatGPT-4}}{$^\triangle$}& 58.9  & \textbf{83.9}  & \textbf{70.3} &\textbf{62.0}&\textbf{91.2}&\textbf{75.2}& 160T (in \& out) & 5.4T/S$^\star$\\
      Few-Shot Prompt 4 \textsc{\textbf{ChatGPT-3.5}}& 53.8$\downarrow$  & 72.8  & 62.6  &57.8&86.2&70.6& 160T (in \& out)& 8.7T/S$^\star$ &1750B\\
      Few-Shot Prompt 4 \textsc{\textbf{Claude3}}& \textbf{60.7}  & 78.6  & 69.1  &59.7&90.5&73.5& 160T (in \& out)\\
      
    \hline
      \textsc{\textbf{OURS-large}}  & \underline{55.1}  & \underline{64.5}  & \underline{59.6}  &\underline{54.7}  &\underline{79.2}&\underline{65.8}& 3.2GB VRAM &147.8T/S  &440M\\
    
\bottomrule
\end{tabular}
\end{table*}


\section{Case Study}
\label{casestudy}
In Table~\ref{case study}, we list successful examples from our model in simple style transfer: formal and informal, positive and negative, and role style transfer: Hu Tao and Noelle. 

For the Comparison in the task of simple style transfer, the three competing models tend to add negative words (e.g., "not") to make the text more negative. 
Additionally, we are surprised that the phrases learned by SETTP  are not present in the training corpus (e.g., "no intention of"). In contrast, the other two models used words in the training corpus.
In the task of role style transfer, we chose two specific characters, Hu Tao and Noelle. Hu Tao is lively and extroverted, while Noelle is meticulous and elegant. 
During the style transfer, our model alters punctuation marks (e.g., "!" "..."), whereas BTTS does not. Additionally, regarding word choice, our transformed words lean more towards  Noelle's character (e.g., "careful"). 


\begin{table*}[!htb]
    \centering
    \caption{Examples of transferred sentences by our method, \textsc{\textbf{BTTS}} and \textsc{\textbf{B-GST}}. Attributes are colored for formality and sentiment transfers in both directions. The deleted words from the input are colored by {\color{red}\underline{red}}, and the inserted words in the output are colored by 
 {\color{blue}\underline{blue}}.}
    \begin{tabular}{p{1cm}|p{7cm}|p{7.5cm}}
    \toprule
     & \textbf{{\color{red}Formal} $\Rightarrow$ {\color{blue}Informal}} & \textbf{{\color{red}Positive} $\Rightarrow$ {\color{blue}Negative}}  \\
    \hline
    \textit{Input} 
    & I {\color{red}\underline{find}} movies {\color{red}\underline{to be a captivating blend}} of art and story.
    & I will {\color{red}\underline{definitely}} come to this restaurant again for a meal.\\
    \hline
    \textsc{\textbf{B-GST}} 
    & I {\color{blue}\underline{feel like}} movies are a {\color{blue}\underline{fun mix}} of art and story. 
    & I have {\color{blue}\underline{no desire}} to dine at this restaurant in the {\color{blue}\underline{future}}.\\
    \hline
    \textsc{\textbf{BTTS}}
    & I {\color{blue}\underline{think}} movies a {\color{blue}\underline{pretty neat mix}} of art and story. 
    & I{\color{blue}\underline{'m unlikely to}} return to this restaurant for {\color{blue}\underline{another}} meal. \\
    \hline
    {\bf SETTP} 
    & I {\color{blue}\underline{reckon}} movies are this {\color{blue}\underline{fun blend}} of art stuff and story.
    & I have {\color{blue}\underline{no intention of}} returning to this restaurant for {\color{blue}\underline{another}} meal.\\
    \hline
    \hline
    & \textbf{{\color{red}Hu Tao} $\Rightarrow$ {\color{blue}Noelle}} & \textbf{{\color{red}Noelle} $\Rightarrow$ {\color{blue}Hu Tao}} \\
    \hline
    \textit{Input}
    & Lemme show you some {\color{red}\underline{fire tricks}}. First... Fire! And then... Whoosh! {\color{red}\underline{Fire butterfly}}! {\color{red}\underline{Be free! }}
    & I cannot {\color{red}\underline{rest on my laurels}}. After all, I am not even {\color{red}\underline{formally}} a knight yet. {\color{red}\underline{Until that day}}, I must work {\color{red}\underline{ even }} {\color{red}\underline{ harder.}}\\
    \hline
    \textsc{\textbf{B-GST}}
    & Allow me to {\color{blue}\underline{demonstrate a bit of my training}}. First, a focused spark... Ignite! Next, {\color{blue}\underline{with precision}}... Whoosh! A butterfly of flames, {\color{blue}\underline{gently}} soaring! {\color{blue}\underline{Away you go, free and guided}}!
    & A {\color{blue}\underline{rest? In my path}}, not a trace! An {\color{blue}\underline{unmarked}} knight in life's whimsical race. I'll {\color{blue}\underline{stride through the day}},  pursuing my dreams with ethereal light{\color{blue}\underline{!}} \\
    \hline
    {\bf SETTP} 
    & Let me share with you a {\color{blue}\underline{technique I've been practicing}}. Firstly, a controlled flame. Ignition. Then, with careful {\color{blue}\underline{guidance}}, a flame takes the {\color{blue}\underline{form of}} a butterfly, dancing freely.
    & ...{\color{blue}\underline{Resting}}? Not my style! A {\color{blue}\underline{formal}} knight, I'm not yet to be. But ah... {\color{blue}\underline{the journey's fun, full of laughter and art}}. Until that day, I'll {\color{blue}\underline{play my spirited part!}} \\
    \bottomrule
    \end{tabular}
    
    \label{case study}
\end{table*}

\section{Expanded Results}
\label{Expanded}
Table~\ref{simple} and Table~\ref{complex} are extensions of Table~\ref{Style Transfer}.

\begin{table*}[!htb]
\centering
  \caption{Comparison of Full Dataset in TST. }\label{simple}
\begin{tabular}{l cc|cc|cc}
\toprule
	\textbf{Dataset} & \multicolumn{2}{c}{YELP \textbf{(CC/ACC/G)}} & \multicolumn{2}{|c}{GYAFC E\&M \textbf{(CC/ACC/G)}} & \multicolumn{2}{|c}{GYAFC F\&R \textbf{(CC/ACC/G)}}  \\
\cmidrule(r){1-1}\cmidrule(r){2-7}
	\textbf{Task Attribute} & \multicolumn{1}{c}{Positive} & \multicolumn{1}{c}{Negative} & \multicolumn{1}{|c}{Formal} & \multicolumn{1}{c}{Informal} & \multicolumn{1}{|c}{Formal} & \multicolumn{1}{c}{Informal}  \\
\midrule[0.5pt]

\textsc{\textbf{CP-G}}                   &  35.3/51.4/42.6  & 36.1/50.8/42.8  & 33.7/68.1/47.9  & 34.5/68.9/48.8  & 37.1/37.4/37.2  & 36.9/36.6/36.7   \\ 
\textsc{\textbf{CP-B}}                   &  36.3/51.2/43.1  & 36.7/51.8/43.6  & 40.6/72.3/54.2  & 39.6/72.7/53.7  & 43.3/43.0/43.1  & 42.3/42.6/42.4   \\ 
\textsc{\textbf{TextSETTR}}              &  44.7/53.8/49.0  & 45.1/55.2/49.9  & 47.5/75.4/59.8  & 46.9/75.8/59.6  & 52.0/51.9/51.9  & 51.4/51.5/51.4   \\ 
\textbf{\textsc{BTTS}}                   &  55.0/53.2/54.1  & 54.4/54.2/54.3  & 53.1/76.2/63.6  & 52.7/75.6/63.1  & 53.9/53.1/53.5  & 52.9/53.7/53.3   \\ \hline
\textsc{\textbf{BART}}$_\textsc{Base}$   &  42.0/54.5/47.8  & 42.2/52.9/47.2  & 31.7/67.1/46.1  & 33.3/67.3/47.3  & 33.8/34.7/34.2  & 35.4/34.5/34.9   \\ 
\textsc{\textbf{BART}}$_\textsc{Large}$  &  49.6/55.1/52.3  & 48.2/55.3/51.6  & 50.7/71.1/60.0  & 50.5/69.7/59.3  & 56.9/56.1/56.5  & 56.7/57.5/57.1   \\ 
\textsc{\textbf{GPT-2}}$_\textsc{Large}$ &  31.8/48.1/39.1  & 32.4/49.1/39.9  & 31.2/55.9/41.8  & 30.2/56.5/41.3  & 32.2/32.0/32.1  & 31.2/31.4/31.3   \\ 
\textsc{\textbf{T5}}$_\textsc{Base}$     &  40.1/54.7/46.8  & 39.7/52.5/45.7  & 36.7/63.1/48.1  & 37.7/62.7/48.6  & 29.5/29.8/29.6  & 30.5/30.2/30.3   \\ 
\textsc{\textbf{T5}}$_\textsc{Large}$    &  52.1/55.8/53.9  & 52.1/57.6/54.8  & 46.0/71.5/57.3  & 45.4/71.5/57.0  & 58.2/57.3/57.7  & 56.4/57.3/56.8   \\ 
\textsc{\textbf{B-GST}}                  &  51.7/56.6/54.1  & 52.5/59.0/55.7  & 54.6/78.9/65.6  & 52.2/79.7/64.5  & 59.8/59.0/59.4  & 57.4/58.2/57.8   \\ 
\textsc{\textbf{Stroytrans}}             &  51.1/53.7/52.4  & 50.3/53.1/51.7  & 51.6/57.6/54.5  & 52.2/56.8/54.5  & 52.2/52.1/52.1  & 52.8/52.9/52.8   \\ \hline
\textsc{\textbf{OURS}}$_\textsc{Base}$   &  42.5/59.2/50.2  & 42.9/60.2/50.8  & 49.9/61.5/55.4  & 48.9/61.9/55.0  & 52.0/51.7/51.8  & 51.0/51.3/51.1   \\ 
\textsc{\textbf{OURS}}$_\textsc{Large}$  &  54.2/62.2/58.1  & 54.8/62.4/58.5  & 54.0/78.9/65.3  & 55.4/79.5/66.4  & 58.2/59.2/58.7  & 59.6/58.6/59.1   \\ 

\bottomrule
\end{tabular}
\end{table*}

\begin{table*}[!htb]
\centering
  \caption{Comparison of Full Dataset in TST. }\label{complex}
\begin{tabular}{l c|c|c|c|c|c}
\toprule
	\textbf{Dataset} & \multicolumn{6}{c}{Genshin \textbf{(CC/ACC/G)}}\\
\cmidrule(r){1-1}\cmidrule(r){2-7}
	\textbf{Task Attribute} & \multicolumn{1}{c}{Hutao} & \multicolumn{1}{c}{Venti} & \multicolumn{1}{c}{Diluc} & \multicolumn{1}{c}{Noelle} & \multicolumn{1}{c}{Mona} & \multicolumn{1}{c}{Xiangling}  \\
 \midrule[0.5pt]

\textsc{\textbf{TextSETTR}}              &   34.7/51.4/42.23  & 31.7/53.0/40.99  & 32.0/52.5/40.99  & 31.0/49.9/39.33  & 34.3/55.1/43.47  & 31.3/53.7/41.00   \\ 
   \textbf{\textsc{BTTS}}                   &     48.8/63.7/55.75  & 47.2/62.1/54.14  & 47.7/63.2/54.91  & 50.3/62.7/56.16  & 45.1/61.9/52.84  & 46.5/62.6/53.95   \\ \hline
   \textsc{\textbf{BART}}$_\textsc{Base}$   &     37.6/48.5/42.70  & 35.2/48.4/41.28  & 38.1/47.1/42.36  & 32.8/47.2/39.35  & 36.1/48.3/41.76  & 39.2/46.1/42.51   \\ 
    \textsc{\textbf{BART}}$_\textsc{Large}$  &     53.9/54.1/54.00  & 55.7/57.1/56.40  & 51.9/56.8/54.29  & 50.7/57.8/54.13  & 55.4/54.5/54.95  & 48.6/57.5/52.86   \\ 
  \textsc{\textbf{GPT-2}}$_\textsc{Large}$ &      33.0/40.5/36.56  & 30.7/42.7/36.21  & 31.1/40.3/35.40  & 31.8/40.9/36.06  & 30.8/41.9/35.92  & 31.6/40.9/35.95   \\ 
   \textsc{\textbf{T5}}$_\textsc{Base}$     &      34.1/42.9/38.25  & 36.6/47.6/41.74  & 38.8/42.5/40.61  & 33.4/43.7/38.20  & 36.0/44.7/40.11  & 37.7/43.2/40.36   \\ 
  \textsc{\textbf{T5}}$_\textsc{Large}$    &       52.4/57.6/54.94  & 53.8/55.1/54.45  & 52.6/52.9/52.75  & 52.3/58.3/55.22  & 53.5/55.7/54.59  & 54.0/54.0/54.00   \\ 
   \textsc{\textbf{B-GST}}                  &      29.8/59.4/42.07  & 29.9/61.6/42.92  & 31.2/59.2/42.98  & 31.1/59.8/43.13  & 30.0/60.8/42.71  & 32.2/59.8/43.88   \\ 
   \textsc{\textbf{Stroytrans}}             &     53.3/45.6/49.30  & 48.6/48.0/48.30  & 53.7/45.1/49.21  & 52.5/50.4/51.44  & 51.5/47.1/49.25  & 53.0/44.0/48.29   \\ 
   \hline
   \textsc{\textbf{OURS}}$_\textsc{Base}$   &     39.3/44.6/41.87  & 40.9/46.9/43.80  & 39.8/46.5/43.02  & 40.3/45.8/42.96  & 41.1/46.8/43.86  & 40.4/46.0/43.11   \\ 
   \textsc{\textbf{OURS}}$_\textsc{Large}$  &      55.8/65.2/60.32  & 53.6/63.8/58.48  & 56.0/65.0/60.33  & 55.4/65.3/60.15  & 54.4/64.1/59.05  & 55.4/63.6/59.36   \\

    \bottomrule
\end{tabular}
\end{table*}

\begin{table*}[!htp]
\caption{Prompts for Evaluation.}
\centering
\footnotesize
\begin{tabular}{lp{13cm}}
\toprule
\multicolumn{1}{c}{\textbf{Baseline}} & \multicolumn{1}{c}{\textbf{Prompt}} \\
\midrule
\textbf{NER} &
{\bf System:}\par
\texttt{[Role][Background][Personality]}\par
{\bf User:}\par
Based on your understanding of \texttt{[Role]}, please rate the following generated content:\par
\texttt{[Generated Text]}\par
Give a continuous rating between 0 (worst) and 100 (best), where 0 means "completely inconsistent with [Role]'s style," and 100 means "this is definitely \texttt{[Role]}'s style, expressed perfectly."\par
Please follow the requirements:\texttt{[Requirements]}\par
{\bf Assistant:}\texttt{[Score][Reasons]}
 \\
\midrule
\textbf{RER} &
\textit{Supervised History:}\par
{\bf System:}\par
\texttt{[Role][Background][Personality]}\par
{\bf User:}\par
Based on your understanding of \texttt{[Role]}, please rate the following generated content:\par
\texttt{[Supervised Text 1]}\par
Give a continuous rating between 0 (worst) and 100 (best), where 0 means "completely inconsistent with [Role]'s style," and 100 means "this is definitely \texttt{[Role]}'s style, expressed perfectly."\par
Please follow the requirements:\texttt{[Requirements]}\par
{\bf Assistant:}\texttt{[Supervised Score 1][Reasons 1]}\par
{\bf User:}\texttt{[Supervised Text 2]}\par
{\bf Assistant:}\texttt{[Supervised Score 2][Reasons 2]}\par
{\bf User:}\texttt{[Supervised Text 3]}\par
{\bf Assistant:}\texttt{[Supervised Score 3][Reasons 3]}\par
{\bf User:}\texttt{[Supervised Text 4]}\par
{\bf Assistant:}\texttt{[Supervised Score 4][Reasons 4]}\par
\textit{Rating:}\par
{\bf User:}\texttt{[Generated Text]}\par
{\bf Assistant:}\texttt{[Score][Reasons]}
 \\
\midrule
\textbf{PCR} &
\textit{Stage 1}\par
{\bf System:}\par
\texttt{[Role][Background][Personality]}\par
{\bf User:}\par
Passage 1:\texttt{[Generated Text]}\par
Passage 2:\texttt{[Golden Text]}\par
Please participate in a vote based on your understanding of \texttt{[Role]}, with the goal of selecting the passage that most closely resembles the style of \texttt{[Role]}. Here are the voting rules:\par
Your voting criteria should include the following five levels:
\textit{Strong Win (Score: 50)}
\textit{Weak Win (Score: 40)}
\textit{Tie (Score: 30)}
\textit{Weak Loss (Score: 20)}
\textit{Strong Loss (Score: 10)}\par
Please follow other requirements:\texttt{[Requirements]} \par
{\bf Assistant:}\texttt{[Score 1][Reasons 1]}\par
\textit{Stage 2}\par
\textbf{\textit{===Exchange the order of Passages===}}\par
{\bf Assistant:}\texttt{[Score 2][Reasons 2]}
 \\
\bottomrule
\end{tabular}
\label{prompt_eval}
\end{table*}

\begin{table*}[!htp]
\caption{Prompts for Comparison with Advanced LLMs.}
\centering
\footnotesize
\begin{tabular}{lp{13cm}}
\toprule
\multicolumn{1}{c}{\textbf{Baseline}} & \multicolumn{1}{c}{\textbf{Prompt}} \\
\midrule
\textbf{Zero-Shot Prompt 1} &
{\bf System:}\texttt{[Role][Background][Personality]}\par
{\bf User:}If you are \texttt{[Role]}, how would you express the following sentence:\par
\texttt{[Original Text]}\par
Please follow the requirements:\par
First, the style has to be very \texttt{[Role]}-like.The contents of the converted sentence must be consistent with the original sentence.\par
{\bf Assistant:}\texttt{[Generated Text]}
 \\
\midrule
\textbf{Few-Shot Prompt 2} &
{\bf System:}\texttt{[Role][Background][Personality]}\par
{\bf User:}This is a dialogue attributed to the character \texttt{[role]}:\par
\texttt{[Golden Text 1]}\par
\texttt{[Golden Text 2]}\par
\texttt{[......]}\par
\texttt{[Golden Text n]}\par
In light of the \texttt{[role]}'s background, distinctive personality traits, and other relevant information, coupled with the given dialogue, you are tasked with precisely emulating the stylistic and tonal aspects of \texttt{[role]}'s speech. The text you are required to mimic is as follows: 
\texttt{[Original Text]}\par
Please follow the requirements:\texttt{[Requirements]}\par
{\bf Assistant:}\texttt{[Generated Text]}
 \\
\midrule
\textbf{Few-Shot Prompt 3} &
{\bf User:}There are two paragraphs from different characters. Rewrite the target paragraph according to the speaking style, personality, mood, word usage and punctuation of the origin paragraph. Pay attention to imitating the speaking style of the first paragraph as much as possible and use different words. Here is an example.\par
Original Paragraph:\texttt{[Original Text 1]}\par
Target Paragraph:\texttt{[Golden Text]}\par
The text you are required to rewrite is as follows: \texttt{[Original Text 2]}\par
Please follow the requirements:\texttt{[Requirements]}\par
{\bf Assistant:}\texttt{[Generated Text]}
 \\
\midrule
\textbf{Few-Shot Prompt 4} &
\textit{Supervised History:}\par
{\bf System:}\texttt{[Role][Background][Personality]}\par
{\bf User:}In light of the \texttt{[role]}'s background, distinctive personality traits, and other relevant information, coupled with the given dialogue, you are tasked with precisely emulating the stylistic and tonal aspects of \texttt{[role]}'s speech. The text you are required to mimic is as follows: \texttt{[Original Text]}\par
Please follow the requirements:\texttt{[Requirements]}\par
{\bf User:}\texttt{[Original Text 1]}\par
{\bf Assistant:}\texttt{[Golden Text 1]}\par
{\bf User:}\texttt{[Original Text 2]}\par
{\bf Assistant:}\texttt{[Golden Text 2]}\par
\texttt{......}\par
{\bf User:}\texttt{[Original Text n]}\par
{\bf Assistant:}\texttt{[Golden Text n]}\par
\textit{Generating:}\par
{\bf User:}\texttt{[Original Text]}\par
{\bf Assistant:}\texttt{[Generated Text]}
 \\

\bottomrule
\end{tabular}

\label{prompt_gen}
\end{table*}

\end{document}